\documentclass[11pt]{article}

\usepackage[margin=1in]{geometry}
\usepackage{amsmath,amssymb}
\usepackage{array}
\usepackage{booktabs}
\usepackage[font=small,labelfont=bf]{caption}
\usepackage{enumitem}
\usepackage{graphicx}
\usepackage{microtype}
\usepackage[numbers]{natbib}
\usepackage{url}
\usepackage{xcolor}
\usepackage{hyperref}

\definecolor{navy}{RGB}{0,56,118}
\hypersetup{
  colorlinks=true,
  linkcolor=navy,
  citecolor=navy,
  urlcolor=navy,
  pdftitle={Topology Hub Optimal Transport},
  pdfauthor={Dylan B. Forde}
}

\newcommand{\one}{\mathbf{1}}
\newcommand{\ReLU}{\operatorname{ReLU}}
\newcommand{\sparsemax}{\operatorname{sparsemax}}
\newcommand{\softmax}{\operatorname{softmax}}

\title{\vspace{-1.2em}\rule{\linewidth}{1.2pt}\\[0.55em]
\textbf{Topology Hub Optimal Transport:}\\
Sparse Capacity Allocation, Exact Tail Differentiation, and Long-Context Efficiency\\[0.35em]
\rule{\linewidth}{0.6pt}}
\author{\textbf{Dylan B.~Forde}}
\date{}

\begin{document}
\maketitle
\vspace{-1.2em}

\begin{abstract}
Long-context models use sparse or structured attention to avoid quadratic cost, yet retrieving a small set of relevant facts embedded among hundreds of thousands of distractors degrades as normalization mass disperses with length. Hub-and-spoke optimal transport (OT) replaces the dense $L \times L$ matrix with a linear-size graph over local windows and $K$ shared hubs balanced by Sinkhorn scaling. Building on block-wise tail refinement and fixed-support transport topology \citep{forde2026blockwise,forde2026topology}, we introduce \emph{Topology Hub Optimal Transport} (THOT): (1) bounded-support capacity gating via simplex $\sparsemax$ ($O(L\log L)$) or sorting-free token-local thresholded $\ReLU$ ($O(L)$), driving excess distractor mass to zero; (2) a three-sweep $1+2$ annealed row-closed Sinkhorn schedule that damps bipartite residuals; (3) an exact rank-$(D+2R+1)$ One-Reference-Tile adjoint in batched split-factor Triton kernels (BF16 Tensor Cores, FP32 potentials); and (4) a chunked exclusive-prefix causal variant with verified zero future leakage. On NVIDIA L4, sparsemax THOT trained at $L=128$ reaches $97.40\%$ zero-shot recall of fixed bindings at $L=262{,}144$ under pure task loss ($\alpha=0$) and $98.83\%$ at $\alpha=0.05$ (matched by $O(L)$ thresholded $\ReLU$ at $\alpha=0.05$, vs.\ $80.60\%$ at $\alpha=0$), while softmax-gated hub OT drops to $3.52\%$. In bidirectional operator microbenchmarks, our kernel runs forward+VJP in $100.1\,\text{ms}$ at $H=8, D=64, L=262\text{K}$ ($2.2\times$ faster with $4.0\times$ smaller compiler temporaries than latent softmax) and $1.34\,\text{ms}$ vs.\ $4.37\,\text{ms}$ for cuDNN FlashAttention at $L=4{,}096$; multi-layer LM runs use the unfused path. Zero-shot WikiText-103 multi-needle retrieval hits $100.00\%$ ($128/128$ queries at $L=65\text{K}$) vs.\ $0.78\%$ for short-trained RoPE FlashAttention, and compact 4-layer WikiText-4K runs show modest causal and masked LM gains.
\end{abstract}

\section{Introduction}
\label{sec:intro}

Scaling attention to sequences of tens or hundreds of thousands of tokens poses two coupled bottlenecks. The computational bottleneck is the quadratic $O(L^2)$ cost of standard attention \citep{vaswani2017attention}: even with IO-aware tiling such as FlashAttention \citep{dao2022flashattention,dao2023flashattention2,shah2024flashattention3}, dense forward-plus-backward passes at $L=262{,}144$ require multiple seconds on a single mid-range GPU. A second bottleneck is normalization dispersion: as irrelevant background tokens accumulate, standard softmax spreads strictly positive probability mass across distractors, degrading retrieval fidelity in long contexts \citep{liu2024lost,hsieh2024ruler,velickovic2025softmax,vasylenko2025longcontext} and making zero-shot extrapolation beyond training lengths sensitive to positional encodings or sink caches \citep{press2022train,peng2024yarn,xiao2024efficient}.

A natural sparse alternative is a \emph{hub-and-spoke} topology---combining a local sliding window of radius $r$ with $K$ shared global hubs \citep{beltagy2020longformer,zaheer2020bigbird}, latent bottleneck arrays \citep{jaegle2021perceiver}, or global register slots \citep{darcet2024vision}. Formulated as an entropic Optimal Transport (OT) problem \citep{cuturi2013sinkhorn,peyre2019computational,tay2020sparse,sander2022sinkformers,shahbazi2025lotformer}, every token and hub receives an input-dependent routing capacity, and Sinkhorn matrix scaling allocates attention mass across $O(L(r+K))$ edges. Unlike unconstrained softmax attention, marginal capacity constraints prevent hubs from collapsing onto a single popular key and enable tokens to regulate their own routing budget---directly targeting the associative-recall fragility observed in sub-quadratic sequence models \citep{arora2024zoology}.

This work builds directly on \citet{forde2026blockwise}, who introduced block-wise differentiable Sinkhorn tail refinement, and \citet{forde2026topology}, who analyzed gradient propagation across fixed sparse transport supports. Scaling fixed-support hub OT to end-to-end training and quarter-million-token evaluation exposed four concrete engineering and modeling bottlenecks:

\begin{enumerate}[leftmargin=1.5em]
  \item \textbf{Softmax capacity dilution ($\Theta(1/(1+\alpha L))$).} Even on an $O(L)$ sparse topology, allocating token capacities with a global softmax assigns strictly positive excess mass to every distractor token. Under a bounded logit margin, cumulative background mass overwhelms the $K$ hubs as sequence length grows, dropping zero-shot associative recall from $98.31\%$ ($L=128$) to $3.52\%$ ($L=262{,}144$).
  \item \textbf{Finite-step parity residuals under bipartite hub routing.} Blocking hub-to-hub shortcuts ($B_{HH} = -20$) isolates two-hop routing through hubs \citep{forde2026blockwise,forde2026topology}, but renders the active subgraph nearly bipartite. In finite Sinkhorn schedules, alternating row and column updates then leave $5\%$--$6\%$ row-marginal residuals unless six or more expensive sweeps are unrolled.
  \item \textbf{Unrolled backward memory and tile-padding overhead.} Naive reverse-mode automatic differentiation through finite Sinkhorn sweeps launches repeated width-$D$ sparse passes and incurs register padding waste, leaving unfused OT over $4\times$ slower than latent softmax attention at $L=262\text{K}$.
  \item \textbf{Non-causality of full-sequence column scaling.} Standard two-sided Sinkhorn column normalization couples future queries into earlier column potentials, violating autoregressive causality.
\end{enumerate}

We address these four issues with \emph{Topology Hub Optimal Transport} (THOT), focusing on sparse capacity allocation, exact finite-sweep differentiation, hardware-efficient bidirectional execution, and a strictly causal chunked variant:

\begin{itemize}[leftmargin=1.5em]
  \item \textbf{Bounded-support capacity gating (\S\ref{sec:gates}).} Replacing global softmax capacity allocation with simplex $\sparsemax$ \citep{martins2016softmax,correia2019adaptively} or a sorting-free token-local thresholded $\ReLU$ gate \citep{wortsman2023replacing} clamps excess distractor capacity to exact zero. Under pure task loss ($\alpha=0$), sparsemax THOT maintains $97.40\%$ zero-shot recall of $R=8$ fixed bindings at $L=262{,}144$ ($98.83\%$ with $\alpha=0.05$); thresholded $\ReLU$ matches $98.83\%$ at $\alpha=0.05$ in $O(L)$ time while dropping to $80.60\%$ at $\alpha=0$.
  \item \textbf{Swept topological annealing and row closure (\S\ref{sec:annealed}).} Starting from a marginal warm start \citep{thornton2023initialization} and local degree offset, one detached aperiodic sweep followed by two differentiated bipartite sweeps and exact row normalization eliminates row-marginal error in three total sweeps and provides a $+1.04$ to $+1.44$ percentage-point refinement (with tighter cross-seed spread at $\alpha=0$) over non-annealed row-closed bipartite OT. We also formalize a prior-preserving semi-relaxed variant (\emph{THOT-SR}) via its exact variational objective.
  \item \textbf{Exact One-Reference-Tile adjoint and split-factor Triton kernels (\S\ref{sec:kernels}, Appendix~\ref{app:details}).} Exploiting diagonal rescaling across finite Sinkhorn iterates, we compress all score gradients across Sweeps 2--3 and row closure into a rank-$(D+2R+1)$ modifier evaluated in \textbf{two} width-$D$ sparse passes over the terminal plan (plus one pass for $\nabla_V$ and four 1D scalar passes), matching exact FP32 differentiation within $4.47\times 10^{-8}$. Batched multi-head Triton kernels \citep{triton2019tillet} with BF16 Tensor Cores and FP32 dual potentials reach $100.1\,\text{ms}$ forward+VJP at $H=8, D=64, L=262\text{K}$ ($2.2\times$ faster with $4.0\times$ smaller compiler temporaries than latent softmax) and $1.34\,\text{ms}$ vs.\ $4.37\,\text{ms}$ for cuDNN FlashAttention at $L=4{,}096$ in operator microbenchmarks.
  \item \textbf{Strictly causal chunked transport and empirical evaluation (\S\ref{sec:causal}--\S\ref{sec:experiments}).} Combining chunk-local write OT, an exclusive prefix scan over past hub summaries, and an exclusive intra-chunk causal prefix log-sum-exp during read OT guarantees exact $0.0$ future dependence (verified by perturbation tests and reverse-mode Jacobians). Causal THOT reaches $90.10\%$ zero-shot recall at $L=262\text{K}$ (with annealed gate warmup), bidirectional sparsemax THOT reaches $100.00\%$ ($128/128$ queries) zero-shot multi-needle retrieval in WikiText-103 prose at $L=65{,}536$ ($\alpha=0$), and compact 4-layer WikiText-4K experiments show modest NLL improvements on both causal and masked language modeling.
\end{itemize}

\section{Background: Failure Modes of Hub-and-Spoke OT}
\label{sec:background}

We first formalize the hub-and-spoke optimal transport attention layer and diagnose why naive formulations fail as distractor length grows or finite-sweep schedules are shortened.

\subsection{Hub-and-Spoke Transport Formulation}

Consider an input sequence of $L$ real tokens augmented with $K \ll L$ shared global hub slots ($N = L + K$ nodes per attention head). Each real token $i \in \{1,\dots,L\}$ connects via an active sparse edge set $\Omega$ to: (i) a circular local window $\{j \in \{1,\dots,L\} : \min(|i-j|, L - |i-j|) \le r\}$, and (ii) all $K$ global hubs; each hub connects to all $N$ nodes (with hub-to-hub edges modulated by $B_{HH}$). Hence $|\Omega| = O(L(2r + 1 + K))$ scales linearly in sequence length $L$, reducing edge count from $O(L^2)$ to $O(L)$.

Given query and key projections $q_i, k_j \in \mathbb{R}^D$, score scale $s_{\text{scale}} = D^{-1/2}$, and a learned $2 \times 2$ node-type bias matrix $B$ indexed by whether $(i,j)$ is real-to-real ($B_{RR}$), real-to-hub ($B_{RH}$), hub-to-real ($B_{HR}$), or hub-to-hub ($B_{HH}$), the edge affinity score is
\begin{equation}
  S_{ij} = \frac{q_i^\top k_j}{\sqrt{D}} + B_{\tau(i),\tau(j)}, \qquad (i,j) \in \Omega.
\end{equation}
Given positive row marginals $a \in \mathbb{R}_{>0}^N$ (emitting budget per query or hub) and column marginals $b \in \mathbb{R}_{>0}^N$ (absorbing budget per key or hub, satisfying $\sum_i a_i = \sum_j b_j = T$ in balanced mode), Sinkhorn scaling \citep{sinkhornknopp1967,cuturi2013sinkhorn,knight2008sinkhorn} iteratively updates log-space dual potentials $u, v \in \mathbb{R}^N$ defining the transport plan $P_{ij} = \exp(S_{ij} + u_i + v_j)$ on $\Omega$:
\begin{align}
  u_i &\leftarrow \ln a_i - \log\sum_{j:(i,j)\in\Omega} \exp(S_{ij} + v_j), \label{eq:sinkhorn-u}\\
  v_j &\leftarrow \ln b_j - \log\sum_{i:(i,j)\in\Omega} \exp(S_{ij} + u_i). \label{eq:sinkhorn-v}
\end{align}
Once row-normalized so that $\sum_j P_{ij} = a_i$, the row-stochastic attention weights are $W_{ij} = P_{ij} / a_i$, aggregating value vectors via $Y = WV = \operatorname{diag}(a)^{-1} (PV)$. Note that a single bidirectional THOT layer executes one hop $Y = WV$ over $\Omega$, updating both token and hub states; consequently, two-hop Source $\to$ Hub $\to$ Query value propagation occurs across two or more stacked Transformer layers (whereas our causal architecture in \S\ref{sec:causal} performs an explicit Write-then-Read transfer inside every layer). Following \citet{forde2026blockwise}, we execute $T_{\text{base}}$ initial sweeps without tracking gradients and backpropagate exactly through a short refinement tail of $T_{\text{tail}}$ sweeps.

\begin{figure}[t]
  \centering
  \includegraphics[width=\linewidth]{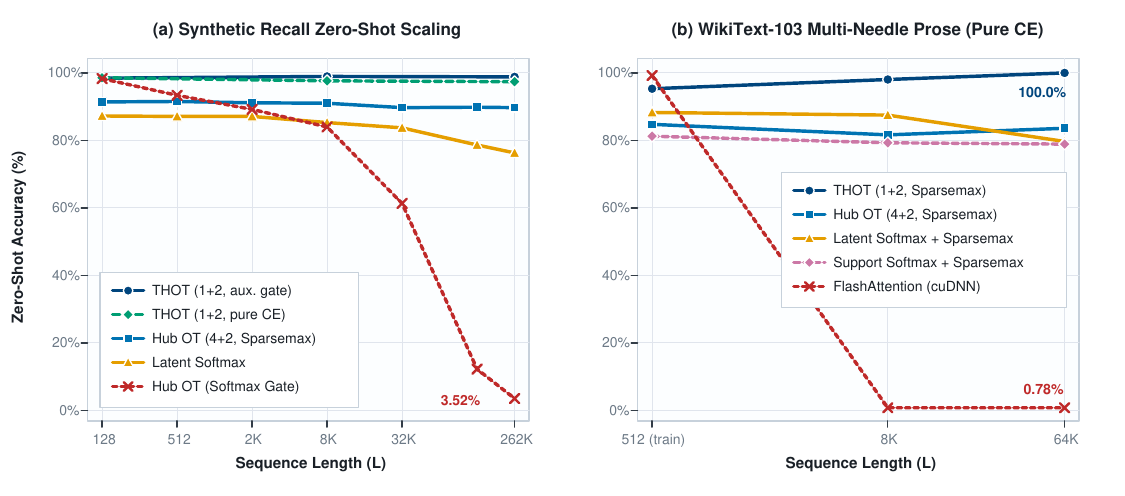}
  \caption{\textbf{Capacity dilution vs.\ THOT from $L=128$ to $L=262{,}144$.}
  \textbf{(a)} On synthetic associative recall ($R=8$ fixed bindings, trained at $L=128$), softmax-gated Hub OT (\emph{Hub OT (Softmax Gate)}, dashed crimson) collapses from $98.31\%$ to $3.52\%$ at $L=262\text{K}$ as distractors absorb $93.4\%$ of excess source capacity. Sparsemax gating with our $1+2$ annealed row-closed schedule (THOT) holds $98.83\%$ with light gate supervision ($\alpha=0.05$) and $97.40\%$ under pure cross-entropy training ($\alpha=0$).
  \textbf{(b)} On zero-shot multi-needle retrieval embedded in WikiText-103 prose (pure cross-entropy, trained at $L=512$), the short-trained RoPE FlashAttention baseline drops from $99.22\%$ to $0.78\%$ beyond its training window, while sparsemax THOT reaches $100.00\%$ ($128/128$ queries, $0.0514$ NLL) at $L=65{,}536$.}
  \label{fig:dilution-and-retrieval}
\end{figure}

\subsection{Failure Mode 1: The \texorpdfstring{$1/(1+\alpha L)$}{1/(1 + alpha L)} Capacity Dilution Law}
\label{sec:dilution}

Suppose a routing gate allocates a total excess capacity budget $M$ across real tokens on top of a uniform local baseline $1/L$, so that $b_j = 1/L + b^{\text{ex}}_j$. Define the \emph{excess filler share} as the fraction of excess source budget landing on irrelevant distractor positions:
\begin{equation}
  \text{Filler Share} = \frac{\sum_{j \in \text{filler}} b^{\text{ex}}_j}{\sum_{j=1}^L b^{\text{ex}}_j}, \qquad b^{\text{ex}}_j = \max\!\left(b_j - \frac{1}{L},\, 0\right).
  \label{eq:filler-share}
\end{equation}
When excess capacities are allocated via a global softmax, $b^{\text{ex}}_j = M \cdot \softmax(g)_j$, suppose $R$ relevant source tokens exceed irrelevant background filler by a logit margin $\Delta$. Because softmax assigns strictly positive weight $\propto \exp(g_{\text{fill}})$ to every distractor, if $\alpha = \exp(-\Delta) > 0$ stays bounded away from zero as $L$ grows, the denominator expands as $R + \alpha(L - R)$. Consequently, the signal fraction of excess routing capacity decays as
\begin{equation}
  \text{Signal Share}(L) = \frac{R}{R + (L - R)e^{-\Delta}} = \Theta\!\left(\frac{1}{1 + \alpha L}\right),
  \label{eq:dilution-law}
\end{equation}
mirroring the softmax dispersion phenomenon analyzed for sharp size generalization by \citet{velickovic2025softmax} and \citet{vasylenko2025longcontext}. From \eqref{eq:dilution-law}, exact-zero sparsity is a \emph{sufficient} solution whenever the learned margin $\Delta$ remains bounded across lengths; in principle, a logit gap $\Delta(L)$ growing as $\Omega(\ln L)$ (for example via length-adaptive temperature \citep{velickovic2025softmax}) could also maintain a nonzero signal share, making length-scaled softmax gating an informative control for future work.

Figure~\ref{fig:dilution-and-retrieval}(a) confirms \eqref{eq:dilution-law} empirically: under fixed-temperature softmax gating, \emph{Hub OT (Softmax Gate)} achieves $98.31\%$ accuracy at $L=128$ ($99.2\%$ excess source share) and $83.98\%$ at $L=8{,}192$ ($67.3\%$ share), before plunging to $12.24\%$ at $L=131\text{K}$ and $3.52\%$ at $L=262{,}144$ ($6.6\%$ signal share, $93.4\%$ excess filler share). Non-OT global-softmax attention with the same structural bias prior (\emph{Support Softmax}) degrades from $12.63\%$ to $0.65\%$. Cross-attending into a fixed $K=16$ bottleneck (\emph{Latent Softmax}, following \citet{jaegle2021perceiver}) resists query-side dilution better, yet key-side softmax normalization still dilutes excess source share from $96.2\%$ down to $1.4\%$, pulling accuracy down from $87.24\%$ to $76.30\%$.

\subsection{Failure Mode 2: Slow Contraction and Parity Residuals Under Bipartite Routing}
\label{sec:oscillation}

Once sparse gating eliminates excess filler capacity, a finite-iteration optimization issue emerges inside the Sinkhorn solver. Prior hub routing \citep{forde2026blockwise} sets a large negative hub-to-hub penalty ($B_{HH} = -20$) so queries route through hubs to source tokens rather than circulating mass among hubs. With distractors reduced to their $1/L$ floor and $B_{HH}$ blocked, however, the active routing mass concentrates on a nearly bipartite subgraph between real tokens ($R$) and hubs ($H$).

To quantify class-level mixing empirically, let $F \in \mathbb{R}^{2\times 2}$ denote the aggregate plan mass between the two node classes, $F_{c,c'} = \sum_{i \in c,\, j \in c'} P_{ij}$ for $c, c' \in \{R, H\}$, and let $Q_{\text{row}} = \operatorname{diag}(F\one)^{-1} F$ and $Q_{\text{col}} = \operatorname{diag}(F^\top\one)^{-1} F^\top$ be the corresponding 1-hop row- and column-normalized class transition matrices. Note that $Q_{\text{row}}$ is distinct from the full-cycle Sinkhorn fixed-point Jacobian on column potentials, $J = \operatorname{diag}(b)^{-1} P^\top \operatorname{diag}(a)^{-1} P$ \citep{knight2008sinkhorn}: $J$ is diagonally similar to a positive-semidefinite Gram matrix and therefore has nonnegative eigenvalues, with $2\times 2$ class quotient $M_{\text{sink}} = Q_{\text{row}} \operatorname{diag}(\gamma_b, 1) Q_{\text{col}}$. However, when $Q_{\text{row}}$ is near-bipartite, its subdominant eigenvalue $\lambda_2(Q_{\text{row}}) = \operatorname{tr}(Q_{\text{row}}) - 1$ lies close to $-1$, which both produces alternating half-step parity imbalance between the $R$ and $H$ partitions and drives the nonnegative two-step quotient eigenvalue of $M_{\text{sink}}$ toward $\lambda_2(Q_{\text{row}})^2 \approx 1$ (slow contraction).

On our $L=262{,}144$ diagnostic (Table~\ref{tab:spectral}), a standard bipartite schedule retains $\lambda_2(Q_{\text{row}}) = -0.9924$ after three sweeps, leaving $5.53\%$ row total-variation (TV) error and only $89.4\%$ query-to-hub flow share. Stopping early at $3$ warm-started sweeps without row closure reaches only $70.57\%$ recall at $L=262\text{K}$ unless extended to $6$ unrolled sweeps ($96.48\%$). Meanwhile, replacing explicit tail differentiation with an uncentered implicit-gradient solver (\emph{Uncentered Implicit-VJP OT}) fails to train ($0.91\%$ at $262\text{K}$).

\section{Topology Hub Optimal Transport}
\label{sec:method}

Topology Hub Optimal Transport (THOT) addresses capacity dilution, finite-step bipartite residuals, backward-pass overhead, and autoregressive causality through four components. Algorithm~1 summarizes the bidirectional forward pass and adjoint, Algorithm~2 specifies the strictly causal chunked variant, and Table~\ref{tab:variants} keeps all evaluated THOT variants explicitly separated.

\subsection{Bounded-Support Capacity Gating: Sparsemax and \texorpdfstring{$O(L)$}{O(L)} Thresholded ReLU}
\label{sec:gates}

To prevent excess capacity from leaking onto distractors under a bounded score margin, we allocate excess budgets $a^{\text{ex}}, b^{\text{ex}}$ with compact support on top of the uniform $1/L$ local baseline:

\paragraph{Simplex $\sparsemax$ gate (\texorpdfstring{$O(L \log L)$}{O(L log L)}).}
Given scalar routing logits $s_{\text{rec}}, s_{\text{src}} \in \mathbb{R}^L$ produced by a lightweight token-wise MLP, we replace $\softmax$ with Euclidean simplex projection $\sparsemax$ \citep{martins2016softmax,correia2019adaptively}:
\begin{equation}
  a_i = \frac{1}{L} + M \cdot \sparsemax(s_{\text{rec}})_i, \qquad
  b_j = \frac{1}{L} + M \cdot \sparsemax(s_{\text{src}})_j,
  \label{eq:sparsemax-gate}
\end{equation}
with total excess mass $M=8.0$ distributed across real tokens and matched across the $K$ hubs ($a^{\text{hub}}_h = b^{\text{hub}}_h = M/K$ for $h=1,\dots,K$; total sequence mass $1 + M$ on real tokens plus $M$ on hubs). Because $\sparsemax$ maps low logits below its data-dependent threshold $\tau(s)$ to exact zero, excess filler share \eqref{eq:filler-share} drops to identically $0.00\%$ across $L=128\dots 262{,}144$ while retaining the $1/L$ baseline on every token for local-window connectivity.

\paragraph{Token-local thresholded ReLU gate (\texorpdfstring{$O(L)$}{O(L)}, sorting-free).}
Global $\sparsemax$ requires sorting across all $L$ tokens, which costs $O(L\log L)$ and couples future tokens through $\tau(s)$ in causal models. Inspired by ReLU attention \citep{wortsman2023replacing}, learned scalar thresholds $\theta_{\text{rec}}, \theta_{\text{src}} \in \mathbb{R}$ (initialized at $-0.25$) yield a sorting-free $O(L)$ gate:
\begin{equation}
  a_i = \frac{1}{L} + M \cdot \frac{\ReLU(s_{\text{rec},i} - \theta_{\text{rec}})}{\sum_{m=1}^L \ReLU(s_{\text{rec},m} - \theta_{\text{rec}}) + \epsilon},
  \label{eq:relu-thresh}
\end{equation}
and symmetrically for $b_j$, or purely token-local unnormalized $a_i = 1/L + c\cdot\ReLU(s_{\text{rec},i} - \theta_{\text{rec}})$ in streaming causal mode. Whenever distractor logits stay below $\theta$, excess filler share \eqref{eq:filler-share} is identically zero without cross-token sorting.

\subsection{Swept Topological Annealing: The \texorpdfstring{$1+2$}{1+2} Row-Closed Schedule}
\label{sec:annealed}

\begin{figure}[t]
  \centering
  \rule{\linewidth}{0.8pt}
  \vspace{-0.35em}
  \begin{flushleft}
  \small
  \hypertarget{alg:thot}{\textbf{Algorithm 1: Bidirectional THOT --- Forward Pass and Exact One-Reference-Tile Adjoint}}
  \end{flushleft}
  \vspace{-0.45em}
  \rule{\linewidth}{0.4pt}
  \begin{minipage}{\linewidth}
  \small
  \setlength{\parskip}{0.25em}
  \textbf{Input:} $(Q, K, V) \in \mathbb{R}^{N \times D}$ ($N=L+K$), node-type biases $B \in \mathbb{R}^{2\times 2}$, local radius $r$, marginals $(a,b) \in \mathbb{R}_{>0}^N$ from \eqref{eq:sparsemax-gate} or \eqref{eq:relu-thresh}.

  \textbf{1.\ Warm start \& degree prior:} Set $v_0 \leftarrow \ln b$ and $B_{RR} \leftarrow B_{RR} + \ln(L/\min(L,2r+1))$.

  \textbf{2.\ Detached aperiodic Sweep 1:} With open hub-to-hub coupling $B_{HH}$ (stop-gradient), run one Sinkhorn cycle \eqref{eq:sinkhorn-u}--\eqref{eq:sinkhorn-v}, center $u_1$ across active rows ($u_1 \leftarrow u^{(1)}_{\text{raw}} - \bar u^{(1)}$, $v_1 \leftarrow v^{(1)}_{\text{raw}} + \bar u^{(1)}$), and detach $v_1 \leftarrow \mathrm{detach}(v_1)$.

  \textbf{3.\ Differentiable bipartite Sweeps 2--3:} Clamp $B_{HH} \leftarrow -20$ and run two centered Sinkhorn cycles starting from $v_1$, producing raw row potentials $(u^{(2)}_{\text{raw}}, u^{(3)}_{\text{raw}})$ and centered pairs $(u_2, v_2), (u_3, v_3)$ (with optional prior-preserving semi-relaxed exponent $\gamma_b=0.92$ on real columns of Sweep 3 for THOT-SR).

  \textbf{4.\ Exact row closure \& row-stochastic read:} Compute
  $u_{\text{out},i} = \ln a_i - \operatorname{LSE}_{j:(i,j)\in\Omega}(S_{ij} + v_{3,j})$,
  terminal plan $P^{\text{out}}_{ij} = \exp(S_{ij} + u_{\text{out},i} + v_{3,j})$ (row sums $a_i$), unnormalized aggregate $Z = P^{\text{out}} V$, and row-stochastic attention output
  $W_{ij} = P^{\text{out}}_{ij}/a_i$, $Y = W V = \operatorname{diag}(a)^{-1} Z$.
  Cache $(Z, v_1, u^{(2)}_{\text{raw}}, u_2, v_2, u^{(3)}_{\text{raw}}, u_3, v_3, u_{\text{out}})$ for backward.

  \textbf{5.\ Exact One-Reference-Tile Adjoint (backward; full recurrence in Appendix~\ref{app:details}):}
  Given $\bar{Y}$, set $\bar{Z}_i = \bar{Y}_i / a_i$ (adding quotient term $-\bar{Y}_i^\top Y_i / a_i$ to $\bar{a}_i$) and compute $\nabla_V = (P^{\text{out}})^\top \bar{Z}$ in one width-$D$ transposed pass. Propagate terminal dual sensitivities through row closure and Sweeps 3--2 via four lightweight 1D scalar sparse matvecs to form the $2R+1=5$ scalar factor columns $(L^{(s)}, R^{(s)}) \in \mathbb{R}^{N \times 5}$ using diagonal rescaling relative to $P^{\text{out}}$. Finally, execute \textbf{two} width-$D$ split-factor sparse passes over the single reference plan $P^{\text{out}}$: one row-major pass for $(\nabla_Q, \nabla_B)$ and one transposed pass for $\nabla_K$.
  \end{minipage}
  \vspace{0.3em}
  \rule{\linewidth}{0.8pt}
\end{figure}

To keep total sweep depth at $3$ while retaining clean bipartite routing and exact row normalization, we use an empirically effective \emph{1+2 annealed row-closed} schedule:
\begin{enumerate}[leftmargin=1.5em]
  \item \textbf{Marginal warm start and degree prior.} We initialize column potentials at the target log-marginals $v_0 = \ln b$ \citep{thornton2023initialization} and add a structural degree prior $\ln(L/d)$ to real-to-real edges $B_{RR}$ (where $d=\min(L,2r+1)$) so local neighborhoods remain balanced against hubs as $L$ scales.
  \item \textbf{Sweep 1 (detached aperiodic warm-start).} Keeping hub-to-hub affinity $B_{HH}$ open rather than clamped to $-20$, we execute one stop-gradient Sinkhorn sweep and center row potentials across active tokens. Opening $B_{HH}$ moves the 1-hop class eigenvalue $\lambda_2(Q_{\text{row}})$ from $-0.99$ to $-0.7451$ on Sweep 1 at $L=262{,}144$ (Table~\ref{tab:spectral}), providing a well-equilibrated initialization before bipartite refinement.
  \item \textbf{Sweeps 2--3 (differentiated bipartite refinement).} Clamping $B_{HH} \leftarrow -20$ suppresses hub-to-hub shortcuts across two differentiable Sinkhorn cycles.
  \item \textbf{Exact row closure and normalization.} Column updates leave $4\%$--$5\%$ row-marginal error at finite sweep count. Performing one final row log-normalization $u_{\text{out},i} = \ln a_i - \operatorname{LSE}_j(S_{ij} + v_{3,j})$ and dividing $P^{\text{out}} V$ by $a_i$ guarantees exact row stochasticity ($\text{row TV} = 0.0000$) and $100.0\%$ Query $\to$ Hub mass utilization.
\end{enumerate}

\paragraph{Prior-preserving semi-relaxed annealing (THOT-SR).}
When many active bindings compete for fewer hubs ($R > K$), relaxing column constraints on real tokens reduces marginal residuals. Standard KL-unbalanced Sinkhorn with unweighted reference kernel $\exp(S_{ij})$ \citep{chizat2018scaling} applies the relaxation exponent $\gamma_b = \rho_b/(1+\rho_b)$ to the entire dual update, $v_j = \gamma_b(\ln b_j - \operatorname{LSE}_i(S_{ij}+u_i))$, which scales down the learned log-capacity prior $\ln b_j$. To preserve the full learned capacity prior $b_j$ while relaxing column matching, we instead minimize the \emph{prior-preserving} semi-relaxed objective using reference measure $K_{ij} = b_j \exp(S_{ij})$:
\begin{equation}
  \min_{P\one = a,\; P \ge 0}\; \operatorname{KL}\!\big(P \,\|\, \exp(S)\operatorname{diag}(b)\big) + \rho_b\,\operatorname{KL}\!\big(P^\top \one \,\|\, b\big).
  \label{eq:prior-semirelax-obj}
\end{equation}
First-order stationarity $\ln P_{ij} = S_{ij} + \ln b_j + u_i - \rho_b \ln(\hat{b}_j / b_j)$ with $v_j := \ln b_j - \rho_b \ln(\hat{b}_j / b_j)$ and $\ln \hat{b}_j = v_j + \operatorname{LSE}_i(S_{ij} + u_i)$ yields $(1 + \rho_b) v_j = (1 + \rho_b)\ln b_j - \rho_b \operatorname{LSE}_i(S_{ij} + u_i)$, hence the exact fixed-point update
\begin{equation}
  v_j = \ln b_j - \gamma_b \operatorname{LSE}_{i:(i,j)\in\Omega}(S_{ij} + u_i), \qquad \gamma_b = \frac{\rho_b}{1 + \rho_b} \in (0, 1],
  \label{eq:prior-semirelax-upd}
\end{equation}
applied on real token columns (keeping $\gamma_b = 1$ on hubs). Applying standard unweighted semi-relaxed updates naively from Sweep 1 (\emph{Naive Semi-Relaxed OT}, $\gamma_b=0.85$) collapses under pure cross-entropy training ($\alpha=0$, $0.13\%$ accuracy; \S\ref{sec:exp-synth}). Preceding prior-preserving relaxation with two balanced sweeps (\emph{THOT-SR}, $\gamma_b=0.92$) preserves $\alpha=0$ stability ($95.57\%$ at $L=262\text{K}$) while cutting six-sweep marginal TV error from $0.0185$ to $0.0102$ (Table~\ref{tab:spectral}). Because the naive ($\gamma_b=0.85$) vs.\ annealed ($\gamma_b=0.92$) comparison changes both the schedule and $\gamma_b$, we also evaluate an intermediate annealed $\gamma_b=0.88$ configuration in Table~\ref{tab:adv-ablations}, which still reaches $88.54\%$ at $\alpha=0$ ($97.53\%$ at $\alpha=0.05$).

\begin{table}[t]
  \centering
  \caption{\textbf{Class-quotient diagnostic and marginal residual audit at $L=262{,}144$.} Here $\lambda_2 = \operatorname{tr}(Q_{\text{row}}) - 1$ is the subdominant eigenvalue of the $2\times 2$ 1-hop Real/Hub class transition matrix (distinct from the PSD two-step Sinkhorn Jacobian), and TV error averages half the $L_1$ row and column deviations from $(a,b)$ at sweep $s_k$. THOT's detached aperiodic Sweep 1 yields $\lambda_2=-0.7451$, exact row stochasticity ($\text{row TV}=0.0000$, $100.0\%$ $Q\!\to\!H$ flow), and the lowest 3-sweep residual among balanced schedules.}
  \label{tab:spectral}
  \setlength{\tabcolsep}{4.5pt}
  \resizebox{\linewidth}{!}{%
  \begin{tabular}{@{}lcccc@{}}
    \toprule
    Schedule & $s_1$ TV ($\lambda_2$) & $s_3$ TV (Row / Col, $\lambda_2$) & $s_6$ TV & $s_3$ Flow ($Q\!\to\!H$ / $H\!\to\!S$) \\
    \midrule
    Standard Bipartite OT & $0.0323$ ($-0.9532$) & $0.0276$ ($0.0553$ / $0.0000$, $-0.9924$) & $0.0263$ & $89.4\%$ / $100.0\%$ \\
    Bipartite OT (3-sweep) & $0.0263$ ($-0.9866$) & $0.0253$ ($0.0506$ / $0.0000$, $-0.9827$) & $0.0231$ & $90.2\%$ / $100.0\%$ \\
    Bipartite OT (3-sweep, RC) & $0.0262$ ($-0.9879$) & $0.0250$ ($0.0000$ / $0.0499$, $-0.9831$) & $0.0227$ & $100.0\%$ / $90.3\%$ \\
    Aperiodic OT (3-sweep, RC) & $0.0760$ ($-0.7451$) & $0.0396$ ($0.0000$ / $0.0792$, $-0.8596$) & $0.0238$ & $100.0\%$ / $84.0\%$ \\
    \textbf{THOT (1+2, Row-Closed)} & $0.0760$ ($-0.7451$) & $\mathbf{0.0209}$ ($\mathbf{0.0000}$ / $0.0419$, $-0.9680$) & $0.0185$ & $\mathbf{100.0\%}$ / $91.7\%$ \\
    \textbf{THOT-SR ($\gamma_b=0.92$)} & $0.0760$ ($-0.7451$) & $\mathbf{0.0201}$ ($\mathbf{0.0000}$ / $0.0402$, $-0.9389$) & $\mathbf{0.0102}$ & $\mathbf{100.0\%}$ / $91.5\%$ \\
    \bottomrule
  \end{tabular}}
\end{table}

\subsection{Hardware-Efficient Execution: Exact One-Reference-Tile Adjoint and Split-Factor Kernels}
\label{sec:kernels}

Naive unrolled reverse-mode differentiation through $R=2$ Sinkhorn sweeps plus terminal row closure launches $2 + 4R + 2 = 12$ width-$D$ score-gradient sparse passes (plus value and vector passes), taking $42.17\,\text{ms}$ forward+VJP and $412.1\,\text{MiB}$ static XLA compiler temporary storage even for a single $D=32, L=262\text{K}$ head. We accelerate bidirectional THOT with two systems optimizations tailored for IO-aware Triton execution \citep{triton2019tillet,flashsinkhorn2026}:

\paragraph{Exact One-Reference-Tile Adjoint.}
Because every intermediate plan $P^{(t)} = \exp(S + u_t + v_t^\top)$ during Sweeps 2--3 acts on the same sparse support $\Omega$ and shares the same edge scores $S$, each intermediate plan is an exact rank-1 diagonal rescaling of the terminal plan $P^{\text{out}}$:
$P^{(t)}_{ij} = P^{\text{out}}_{ij} \exp(u_{t,i} - u_{\text{out},i}) \exp(v_{t,j} - v_{3,j})$.
By caching the forward tail potentials $(v_1, u^{(2)}_{\text{raw}}, u_2, v_2, u^{(3)}_{\text{raw}}, u_3, v_3, u_{\text{out}})$---avoiding all five recomputation log-sum-exp launches---the exact reverse recurrence over Sweeps 3--2 and terminal row closure computes all marginal gradients $(\nabla_a, \nabla_b)$ and constructs $2R+1 = 5$ scalar factor columns $(L^{(s)}, R^{(s)}) \in \mathbb{R}^{N \times 5}$ using four lightweight 1D scalar sparse matvecs (no width-$D$ payloads). Combining these with the width-$D$ terminal factors $(\bar{Z}, V)$ yields the exact rank-$(D+2R+1)$ score adjoint (derived in full in Appendix~\ref{app:details}):
\begin{equation}
  \bar{S}_{ij} = P^{\text{out}}_{ij} \left( \bar{Z}_i^\top V_j + \sum_{m=1}^{2R+1} L^{(s)}_{i,m}\, R^{(s)}_{j,m} \right).
  \label{eq:one-ref-score-main}
\end{equation}
All parameter score gradients $(\nabla_Q, \nabla_K, \nabla_B)$ then execute in \textbf{two} width-$D$ sparse passes over the single reference plan $P^{\text{out}}$ (one row-major pass for $\nabla_Q$ and $\nabla_B$, one transposed pass for $\nabla_K$), alongside one width-$D$ transposed pass for $\nabla_V$. On a single $D=32, L=262\text{K}$ head, this alone cuts forward+VJP latency by $1.58\times$ ($42.17 \to 26.64\,\text{ms}$) and shrinks static compiler temporaries by $2.86\times$ ($412.1 \to 144.0\,\text{MiB}$).

\paragraph{Split-Factor Multi-Head Triton Tiling with BF16 Tensor Cores.}
We further remove four multi-head GPU bottlenecks in our custom Triton kernels:
\begin{enumerate}[leftmargin=1.5em]
  \item \textbf{Split feature and scalar-factor accumulators:} Concatenating the width-$D$ factors $(\bar{Z}, V)$ and width-$(2R+1)$ scalar factors $(L^{(s)}, R^{(s)})$ into a single tile forces power-of-two padding up to width $128$ at $D=64$. Keeping them as two separate SRAM dot products---width $D$ and width $2R+1=5$ (padded to $16$)---halves register footprint.
  \item \textbf{Native head-batch grid scheduling:} Dispatching across heads and batch items directly in the GPU thread-block grid eliminates per-head vectorization wrappers and removes padding/concatenation HBM copies on every half-sweep via predicated register masking.
  \item \textbf{Dedicated 1D scalar matvecs and fused bias gradients:} The four reverse-recurrence scalar passes skip vector payload tiling altogether, and the $2\times 2$ node-type bias gradient $\nabla_B$ accumulates directly in registers during the row-major $\nabla_Q$ pass (with $\nabla_{B_{HH}} = 0$ on the bipartite tail).
  \item \textbf{Mixed-precision BF16 Tensor Cores with FP32 duals:} Matrix multiplies execute on BF16 Tensor Cores with FP32 accumulation, while potentials $(u, v)$ and log-sum-exp states stay strictly FP32---analogous to mixed-precision FlashAttention \citep{dao2023flashattention2,shah2024flashattention3}. Against the unfused FP32 reference implementation across $(Y, \nabla_Q, \nabla_K, \nabla_V, \nabla_a, \nabla_b, \nabla_B)$ on $L \in \{128, 256\}$ ($H \in \{2,4\}, D \in \{32,64\}$), worst-case maximum absolute error is $4.47\times 10^{-8}$ (relative $L_2 \le 5.02\times 10^{-7}$) in FP32 mode and $3.27\times 10^{-4}$ in BF16 Tensor-Core mode (and independently matches pure-JAX unrolled dense autodiff within $\le 6.56\times 10^{-7}$ in FP32).
\end{enumerate}

\subsection{Strictly Causal Autoregressive Chunked Transport}
\label{sec:causal}

\begin{figure}[t]
  \centering
  \rule{\linewidth}{0.8pt}
  \vspace{-0.35em}
  \begin{flushleft}
  \small
  \hypertarget{alg:causal-thot}{\textbf{Algorithm 2: Strictly Causal Chunked THOT (Single Attention Head)}}
  \end{flushleft}
  \vspace{-0.45em}
  \rule{\linewidth}{0.4pt}
  \begin{minipage}{\linewidth}
  \small
  \setlength{\parskip}{0.25em}
  \textbf{Input:} Token tensors $(Q, K, V) \in \mathbb{R}^{L \times D}$ (optional position-invariant write keys $K^{\text{w}}$), token-local capacities $(a, b) \in \mathbb{R}_{>0}^L$, hub slots $(Q^{\text{hub}}_0, K^{\text{hub}}_0, V^{\text{hub}}_0) \in \mathbb{R}^{K \times D}$, bias $B \in \mathbb{R}^{2\times 2}$, chunk size $C=32$, local radius $r$. Partition sequence into $N_c = \lceil L/C \rceil$ chunks.

  \textbf{1.\ Chunk-local Write OT ($K$ hubs $\leftarrow$ $C$ tokens per chunk $c$):}
  With chunk score matrix $S^{\text{w}}_c = s_{\text{scale}} Q^{\text{hub}}_0 (K^{\text{w}}_c)^\top + B_{HR}$ and hub capacity cap $\kappa_c = \max(\frac{1}{K}\sum_{j\in c} b_{c,j},\, \max_{j\in c} b_{c,j})$, run 1 detached base sweep at scale $s_{\text{scale}}/\varepsilon_f$ ($\varepsilon_f=1.5$) followed by 2 differentiated tail sweeps and final column closure:
  \[
    u^{\text{w}}_{c,k} \leftarrow -\gamma_w \ReLU\!\big(\operatorname{LSE}_{j \in c}(S^{\text{w}}_{c,k,j} + v^{\text{w}}_{c,j}) - \ln \kappa_c\big),\quad
    v^{\text{w}}_{c,j} \leftarrow \ln b_{c,j} - \operatorname{LSE}_{k=1}^{K}(S^{\text{w}}_{c,k,j} + u^{\text{w}}_{c,k}),
  \]
  with $\gamma_w = 0.70$, yielding $P^{\text{write}}_c \in \mathbb{R}^{K\times C}$ satisfying $\sum_k P^{\text{write}}_{c,k,j} = b_{c,j}$. Compute chunk moments $W^{(V)}_c = P^{\text{write}}_c V_c$, $W^{(K)}_c = P^{\text{write}}_c K^{\text{w}}_c$, and hub mass $z_c = P^{\text{write}}_c \one_C$.

  \textbf{2.\ Exclusive prefix accumulation across past chunks ($c' < c$):}
  Compute exclusive prefix sums $(M^{(V)}_{<c}, M^{(K)}_{<c}, z_{<c}) = \sum_{c' < c} (W^{(V)}_{c'}, W^{(K)}_{c'}, z_{c'})$ (zero for $c=0$), and form normalized past hub summaries with prior weight $m_0 = 0.05$ and residual key blend $\beta = 0.80$:
  \[
    V^{\text{hub}}_{<c} = \frac{m_0 V^{\text{hub}}_0 + M^{(V)}_{<c}}{m_0 + z_{<c}}, \qquad
    K^{\text{hub}}_{<c} = (1-\beta) K^{\text{hub}}_0 + \beta\,\frac{m_0 K^{\text{hub}}_0 + M^{(K)}_{<c}}{m_0 + z_{<c}},
  \]
  and bounded past hub capacity $b^{\text{hub}}_{<c,k} = \min(\sum_h z_{<c,h},\, M_{\text{hub}})\,\frac{z_{<c,k}}{\sum_h z_{<c,h}}$ ($M_{\text{hub}}=8.0$).

  \textbf{3.\ Strictly causal Read OT ($C$ queries $\leftarrow$ $(r+C)$ local past + $K$ past hubs):}
  For chunk $c$, each query $i \in \{0,\dots,C-1\}$ (global index $t=cC+i$) connects to $\Omega_{c,i} = \{j : \max(0, t-r) \le j \le t\} \cup \{H_{<c,k} : c > 0,\, z_{<c,k} > 0\}$ with source capacities $b^{\text{read}}_c = [b^{\text{loc}}_c;\, b^{\text{hub}}_{<c}]$. Initialize $v^{\text{r}}_{c,i,s} = \ln b^{\text{read}}_{c,s}$ and run 1 detached base sweep ($s_{\text{scale}}/\varepsilon_f$), 2 differentiated tail sweeps, and final row closure, where every column reduction uses an \emph{exclusive causal prefix log-sum-exp} strictly over earlier queries $0 \le i' < i$ within chunk $c$:
  \[
    u^{\text{r}}_{c,i} \leftarrow \ln a_{c,i} - \operatorname{LSE}_{s \in \Omega_{c,i}}(S^{\text{read}}_{c,i,s} + v^{\text{r}}_{c,i,s}),\qquad
    v^{\text{r}}_{c,i,s} \leftarrow \ln b^{\text{read}}_{c,s} - \gamma_r \operatorname{LSE}_{\substack{0 \le i' < i \\ s \in \Omega_{c,i'}}}(S^{\text{read}}_{c,i',s} + u^{\text{r}}_{c,i'}),
  \]
  with $\gamma_r = 0.25$. Return $Y_{c,i} = a_{c,i}^{-1} \sum_{s \in \Omega_{c,i}} \exp(S^{\text{read}}_{c,i,s} + u^{\text{r,out}}_{c,i} + v^{\text{r}}_{c,i,s}) V^{\text{read}}_{c,s}$.
  \end{minipage}
  \vspace{0.3em}
  \rule{\linewidth}{0.8pt}
\end{figure}

Full-sequence Sinkhorn scaling couples future keys and queries into earlier potentials via \eqref{eq:sinkhorn-v}. To support linear-time autoregressive modeling alongside causal linear attention, selective state spaces, and slot-memory architectures \citep{katharopoulos2020transformers,gu2024mamba,yang2024gla,zhang2024gsa}, we partition sequences into chunks of size $C=32$ and replace both global sorting and intra-chunk column coupling with strictly causal operations, specified completely in \hyperlink{alg:causal-thot}{Algorithm~2}.

\paragraph{Causal dependency guarantee.}
Consider any output position $t = cC + i$. (i) Its capacities $(a_t, b_t) = 1/L + \ReLU(s_t - \theta)$ depend strictly on token $t$. (ii) Its past hub summaries $(K^{\text{hub}}_{<c}, V^{\text{hub}}_{<c}, b^{\text{hub}}_{<c})$ aggregate only completed chunks $c' < c$, all of whose tokens satisfy $t' < cC \le t$. (iii) Its local candidate set $\Omega_{c,i}$ contains only positions $j \le t$. (iv) Crucially, during read-side Sinkhorn sweeps inside chunk $c$, the column potential $v^{\text{r}}_{c,i,s}$ reduces strictly over earlier queries $0 \le i' < i$ via an exclusive prefix log-sum-exp, preventing later queries $i' > i$ in the same chunk from altering $v^{\text{r}}_{c,i,s}$. By induction over the finite sweeps, no intermediate state entering $Y_t$ depends on any input at $t' > t$. Both adversarial future-perturbation checks across chunk boundaries and exact reverse-mode Jacobians ($\partial Y_t / \partial \{K_{t'}, V_{t'}, s^{\text{src}}_{t'}\}$ for $t < t'$) verify a maximum leakage of identically \textbf{0.0}.

\subsection{Summary of Evaluated THOT Variants}
\label{sec:variants}

Because our empirical study evaluates four distinct THOT configurations with different complexity, causality, and supervision properties, Table~\ref{tab:variants} summarizes their exact boundaries so claims are never conflated across variants.

\begin{table}[htbp]
  \centering
  \caption{\textbf{Summary of evaluated THOT configurations.} No single configuration simultaneously attains $97.40\%$ pure-$\alpha=0$ recall, sorting-free $O(L)$ gating, and causality; each row reports the exact observed trade-off at $L=262{,}144$.}
  \label{tab:variants}
  \setlength{\tabcolsep}{4pt}
  \resizebox{\linewidth}{!}{%
  \begin{tabular}{@{}lccccc@{}}
    \toprule
    Variant & Gate Cost & Causality & Gate Supervision ($\alpha$) & $L=262\text{K}$ Recall & Primary Trade-off \\
    \midrule
    THOT (1+2, Sparsemax) & $O(L \log L)$ & Bidirectional & $\alpha=0$ / $\alpha=0.05$ & $97.40\%$ / $98.83\%$ & Strongest $\alpha=0$ recall; requires global sort \\
    THOT (1+2, ReLU-Thresh) & $O(L)$ & Bidirectional & $\alpha=0$ / $\alpha=0.05$ & $80.60\%$ / $98.83\%$ & Sorting-free; needs light $\alpha$ for $>98\%$ \\
    THOT-SR ($\gamma_b=0.92$) & $O(L \log L)$ & Bidirectional & $\alpha=0$ / $\alpha=0.05$ & $95.57\%$ / $97.40\%$ & Lowest 6-sweep TV \& best $R=32$ overload \\
    Causal THOT (Alg.~2) & $O(L)$ ($O(1)$/tok) & Strictly causal & Annealed $1\!\to\!0$ ($256$ steps) & $90.10\%$ & Exact $0.0$ future leak; unfused chunk path \\
    \bottomrule
  \end{tabular}}
\end{table}

\section{Experiments}
\label{sec:experiments}

All experiments were conducted on NVIDIA L4 GPUs using JAX (0.11.1, NumPy 2.4.6) under matched training budgets, fixed random seeds, and deterministic evaluation splits. For all length-extrapolation benchmarks, models are trained exclusively on short sequences ($L=128$ or $L=512$) and evaluated zero-shot without length fine-tuning out to $L=65{,}536$ or $L=262{,}144$. Individual per-seed outcomes and key experimental configurations appear in Appendix~\ref{app:details}.

\subsection{Synthetic Long-Context Extrapolation, Pure \texorpdfstring{$\alpha=0$}{alpha=0} Training, and Ablations}
\label{sec:exp-synth}

\paragraph{Experimental setup.}
We evaluate retrieval of a \emph{fixed sparse set} of $R=8$ key-value bindings embedded among growing distractor sequences using a 2-layer MoE Transformer ($D=32$, 4 experts, $K=16$ hubs, local radius $r=4$, $Q=8$ queries over $512$ distinct keys and $256$ values). Each model trains for $1{,}536$ updates at $L=128$ across three seeds ($1711, 2917, 4129$) and is evaluated zero-shot on 32 held-out sequences per seed (256 queries/seed; 768 queries across 3 seeds) for length scaling ($L \in \{128, 8{,}192, 262{,}144\}$), $L=8{,}192$ key-group collision stress ($C \in \{0,16,32,64\}$ competing distractors sharing a key group), and $L=512$ hub-capacity overload ($R=Q \in \{8,16,32\}$ against $K=16$ hubs). Auxiliary gate supervision weight $\alpha \in \{0.05, 0\}$ tests whether routing emerges purely from task cross-entropy ($\alpha=0$).

\begin{table}[t]
  \centering
  \caption{\textbf{Zero-shot distractor extrapolation ($L=128 \to 262\text{K}$, $R=8$ fixed bindings), key-group collisions ($L=8\text{K}$), and capacity overload ($L=512$)} across 3 seeds (768 queries total per condition). Bounded-support gating eliminates softmax capacity collapse ($3.52\%$), while row-closed THOT reaches $98.83\%$ ($\alpha=0.05$) and $97.40\%$ ($\alpha=0$) at $L=262{,}144$.}
  \label{tab:synth-main}
  \setlength{\tabcolsep}{3.8pt}
  \resizebox{\linewidth}{!}{%
  \begin{tabular}{@{}lcccccc@{}}
    \toprule
    Method & $\alpha$ & $L=128$ & $L=8{,}192$ & $L=262{,}144$ & Coll.\ $8\text{K}$ (C0/16/32/64) & Cap.\ $512$ (R8/16/32) \\
    \midrule
    Hub OT (Softmax Gate) & $0.05$ & $98.31\%$ & $83.98\%$ & $3.52\%$ & --- & --- \\
    Support Softmax & $0.05$ & $12.63\%$ & $4.17\%$ & $0.65\%$ & --- & --- \\
    Support Softmax + Sparsemax & $0.05$ & $38.15\%$ & $38.02\%$ & $38.02\%$ & $38.80 / 13.02 / 7.29 / 2.21$ & $36.85 / 18.49 / 8.76$ \\
    Latent Softmax & $0.05$ & $87.24\%$ & $85.29\%$ & $76.30\%$ & --- & --- \\
    Latent Softmax + Sparsemax & $0.05$ & $84.77\%$ & $81.77\%$ & $69.79\%$ & $81.38 / 38.67 / 19.66 / 9.24$ & $84.64 / 64.06 / 37.92$ \\
    Hub OT (4+2, Sparsemax) & $0.05$ & $91.41\%$ & $91.02\%$ & $89.71\%$ & $90.76 / 37.11 / 16.02 / 6.90$ & $89.71 / 62.96 / 26.24$ \\
    Bipartite OT (3-sweep, RC) & $0.05$ & $97.40\%$ & $97.66\%$ & $97.79\%$ & $96.88 / 39.19 / 22.53 / 10.03$ & $97.92 / 75.98 / 36.07$ \\
    \textbf{THOT (1+2, Sparsemax)} & $\mathbf{0.05}$ & $\mathbf{98.44\%}$ & $\mathbf{98.96\%}$ & $\mathbf{98.83\%}$ & $\mathbf{97.92} / 42.06 / 20.44 / 9.77$ & $97.01 / \mathbf{78.12} / \mathbf{37.66}$ \\
    \midrule
    Bipartite OT (3-sweep, RC) & $0.00$ & $98.18\%$ & $95.57\%$ & $95.96\%$ & $95.18 / \mathbf{44.53} / \mathbf{24.48} / \mathbf{12.76}$ & $96.22 / 75.00 / 36.52$ \\
    \textbf{THOT (1+2, Sparsemax)} & $\mathbf{0.00}$ & $\mathbf{98.57\%}$ & $\mathbf{97.66\%}$ & $\mathbf{97.40\%}$ & $\mathbf{97.53} / 36.33 / 18.88 / 9.24$ & $\mathbf{97.92} / \mathbf{79.95} / \mathbf{38.41}$ \\
    \bottomrule
  \end{tabular}}
\end{table}

\paragraph{Findings (Tables~\ref{tab:synth-main}--\ref{tab:adv-ablations}).}
The primary jump in long-context recall comes from sparse capacity gating and row closure: replacing softmax gating with $\sparsemax$ lifts $L=262\text{K}$ accuracy from $3.52\%$ to $89.71\%$ even on the legacy $4+2$ schedule, and adding exact row closure (\emph{Bipartite OT (3-sweep, RC)}) raises recall to $97.79\%$ ($\alpha=0.05$) and $95.96\%$ ($\alpha=0$). Adding $1+2$ topological annealing (\emph{THOT (1+2, Sparsemax)}) provides a further $+1.04$ to $+1.44$ percentage-point refinement on zero-shot $262\text{K}$ extrapolation ($98.83\%$ at $\alpha=0.05$, outperforming Bipartite RC on all three seeds; $97.40\%$ at $\alpha=0$ with tightly clustered seeds $(96.88, 97.27, 98.05)\%$ vs.\ $(99.61, 99.22, 89.06)\%$ for Bipartite RC; Table~\ref{tab:per-seed}). Notably, this schedule exhibits a clear trade-off under $\alpha=0$: non-annealed Bipartite RC achieves stronger accuracy across all three nonzero collision levels ($C \in \{16, 32, 64\}$), whereas annealed THOT leads on clean extrapolation and hub capacity overload ($R \in \{8, 16, 32\}$). Table~\ref{tab:adv-ablations} summarizes our gate, semi-relaxed, multi-head, and negative-control ablations:
\begin{itemize}[leftmargin=1.5em]
  \item \textbf{Sorting-free $O(L)$ gating:} \emph{THOT (1+2, ReLU-Thresh)} matches $\sparsemax$ at $\alpha=0.05$ ($98.83\%$ at $L=262\text{K}$, $80.66\%$ at $R=16$) with zero cross-token sorting, lowering gate-only forward latency from $0.355\,\text{ms}$ to $0.220\,\text{ms}$ at $L=8\text{K}$ and from $0.709\,\text{ms}$ to $0.596\,\text{ms}$ at $L=262\text{K}$. Without gate supervision ($\alpha=0$), however, ReLU thresholding drops to $80.60\%$ at $L=262\text{K}$, whereas $\sparsemax$ retains $97.40\%$.
  \item \textbf{Prior-preserving semi-relaxed annealing:} Naive unweighted semi-relaxed OT ($\gamma_b=0.85$) collapses completely at $\alpha=0$ ($0.13\%$ accuracy). Preceding prior-preserving relaxation with two balanced sweeps (\emph{THOT-SR}) restores $\alpha=0$ stability at both $\gamma_b=0.92$ ($95.57\%$ at $262\text{K}$; strongest $R=32$ overload at $39.91\%$) and $\gamma_b=0.88$ ($88.54\%$ at $\alpha=0$, $97.53\%$ at $\alpha=0.05$).
  \item \textbf{Multi-head trade-off and negative controls:} Parameter-matched 4-head THOT ($H=4, d_h=8$, $56{,}651 \le 56{,}688$ active parameters) retains $95.96\%$ ($\alpha=0.05$) and $94.14\%$ ($\alpha=0$) at $L=262\text{K}$, though narrower head dimension slightly lowers overloaded $R=32$ recall ($34.31\%$ vs.\ $38.41\%$). Across all variants, severe overload ($R=32$ bindings against $K=16$ hubs) remains bounded near $38\%$--$40\%$, confirming that our $262\text{K}$ result measures distractor tolerance for a fixed sparse set rather than unbounded hub capacity. Uncentered implicit differentiation fails to train ($0.91\%$ at $262\text{K}$).
\end{itemize}

\begin{table}[t]
  \centering
  \caption{\textbf{Gate complexity, semi-relaxed annealing, multi-head scaling, and negative controls} across 3 seeds (768 queries per condition). Naive semi-relaxed OT and uncentered implicit differentiation fail during training, whereas 2-stage balanced-to-semirelaxed annealing (THOT-SR) and $O(L)$ thresholded ReLU (with $\alpha=0.05$) sustain high accuracy at $L=262\text{K}$.}
  \label{tab:adv-ablations}
  \setlength{\tabcolsep}{4pt}
  \resizebox{\linewidth}{!}{%
  \begin{tabular}{@{}lcccc@{}}
    \toprule
    Variant & $\alpha$ & $L=128$ / $8\text{K}$ / $262\text{K}$ & Coll.\ $8\text{K}$ (C0/16/32/64) & Cap.\ $512$ (R8/16/32) \\
    \midrule
    THOT (1+2, ReLU-Thresh, $O(L)$) & $0.05$ & $\mathbf{99.09 / 98.96 / 98.83}$ & $98.18 / 40.10 / 20.83 / 8.98$ & $98.05 / \mathbf{80.66} / 39.81$ \\
    THOT (1+2, ReLU-Unnorm, $O(L)$) & $0.05$ & $97.92 / 96.88 / 94.79$ & $96.74 / 42.58 / 18.62 / 7.55$ & $97.66 / 78.32 / \mathbf{40.20}$ \\
    THOT-SR ($\gamma_b=0.92$, Sparsemax) & $0.05$ & $97.79 / 97.53 / 97.40$ & $\mathbf{98.57} / 42.06 / 23.57 / \mathbf{11.33}$ & $97.53 / 78.84 / 38.90$ \\
    THOT-SR ($\gamma_b=0.88$, Sparsemax) & $0.05$ & $97.66 / 97.66 / 97.53$ & $96.88 / \mathbf{45.05} / \mathbf{23.83} / 10.68$ & $97.53 / 77.73 / 39.06$ \\
    THOT (4-Head, $d_h=8$, Sparsemax) & $0.05$ & $96.74 / 96.74 / 95.96$ & $95.57 / 38.93 / 18.49 / 6.51$ & $95.31 / 73.63 / 34.31$ \\
    \midrule
    THOT (1+2, ReLU-Thresh, $O(L)$) & $0.00$ & $97.92 / 93.49 / 80.60$ & $93.62 / 35.16 / 18.10 / 7.03$ & $96.74 / 77.67 / 37.57$ \\
    \textbf{THOT-SR ($\gamma_b=0.92$, Sparsemax)} & $\mathbf{0.00}$ & $\mathbf{98.44 / 97.01 / 95.57}$ & $\mathbf{95.70} / \mathbf{40.23} / \mathbf{22.53} / \mathbf{10.03}$ & $\mathbf{97.27} / \mathbf{76.89} / \mathbf{39.91}$ \\
    THOT-SR ($\gamma_b=0.88$, Sparsemax) & $0.00$ & $91.93 / 89.19 / 88.54$ & $90.23 / 36.20 / 20.18 / 9.24$ & $91.28 / 72.14 / 36.52$ \\
    THOT (4-Head, $d_h=8$, Sparsemax) & $0.00$ & $97.01 / 95.44 / 94.14$ & $95.05 / 36.98 / 15.76 / 6.77$ & $96.09 / 72.98 / 34.67$ \\
    \midrule
    \textit{Control:} Naive Semi-Relaxed OT ($\gamma_b=0.85$) & $0.00$ & $0.13 / 0.13 / 0.13$ & $0.65 / 0.65 / 0.65 / 0.65$ & $0.26 / 0.33 / 0.33$ \\
    \textit{Control:} Uncentered Implicit-VJP OT & $0.05$ & $0.65 / 0.52 / 0.91$ & --- & --- \\
    \bottomrule
  \end{tabular}}
\end{table}

\subsection{GPU Operator Microbenchmarks and Compiler Memory}
\label{sec:exp-sys}

\begin{figure}[t]
  \centering
  \includegraphics[width=\linewidth]{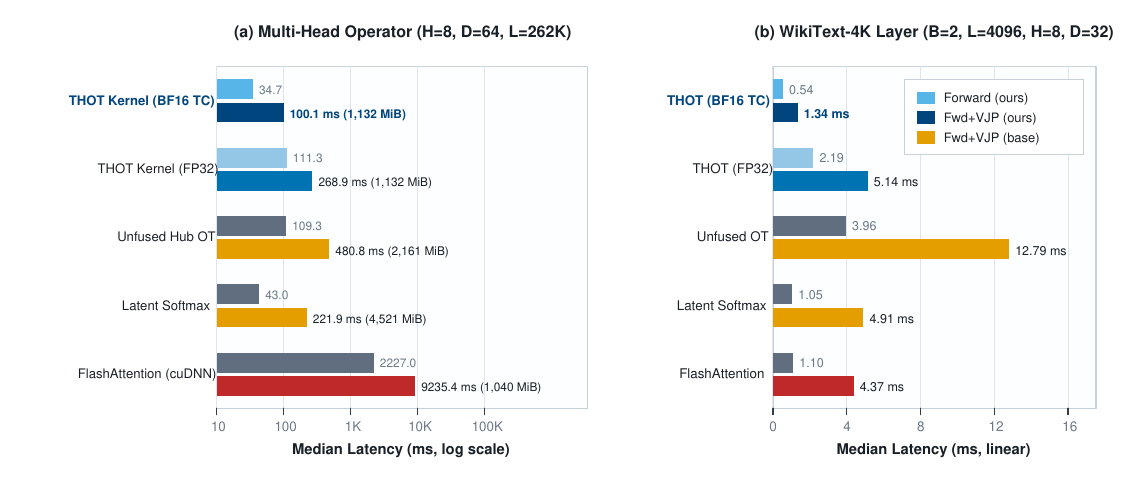}
  \caption{\textbf{Bidirectional GPU operator latency and static XLA compiler temporary memory on NVIDIA L4.}
  \textbf{(a)} At multi-head transformer geometry ($H=8, D=64, L=262{,}144$), our batched split-factor BF16 Tensor-Core kernel completes forward+VJP in $100.1\,\text{ms}$ ($1{,}132\,\text{MiB}$ static compiler temp), outperforming Unfused Hub OT by $4.8\times$, Latent Softmax by $2.2\times$ ($4.0\times$ smaller compiler temporaries than Latent Softmax), and dense cuDNN FlashAttention by $92\times$.
  \textbf{(b)} On a bidirectional WikiText-4K layer geometry ($B=2, L=4{,}096, H=8, D=32, r=32$), THOT (BF16 TC) runs forward+VJP $3.3\times$ faster than cuDNN FlashAttention ($1.34\,\text{ms}$ vs.\ $4.37\,\text{ms}$).}
  \label{fig:systems-pareto}
\end{figure}

\paragraph{Benchmark protocol and scope.}
Table~\ref{tab:systems} and Figure~\ref{fig:systems-pareto} benchmark the standalone \emph{bidirectional} attention operators on NVIDIA L4 (JAX 0.11.1, jaxlib 0.11.1, NumPy 2.4.6), reporting the median of five device-synchronized executions after JIT compilation warmup. Every operator receives identical BF16 $(Q, K, V)$ inputs (plus FP32 marginals and biases where applicable). \emph{FlashAttention (cuDNN)} executes cuDNN BF16 attention; \emph{Unfused Hub OT}, \emph{Latent Softmax}, and \emph{THOT Kernel (FP32)} execute FP32 matmuls and reductions; \emph{THOT Kernel (BF16 TC)} executes BF16 Tensor-Core matmuls with FP32 accumulation and strictly FP32 dual potentials. Timings isolate the attention operator itself (plus hub-RMSNorm/content-slice assembly in the WikiText-4K layer row), excluding QKV projections, capacity MLP/sorting, and FFN/MoE blocks. Importantly, Table~\ref{tab:systems} evaluates the bidirectional kernel, \emph{not} the causal chunked scan (\S\ref{sec:causal}), and these kernels have not yet been wired into the multi-layer WikiText-4K training loops (where full-model step times remain slower than dense FlashAttention: $174.15$ vs.\ $136.32\,\text{ms}$ in MLM and $211.41$ vs.\ $126.45\,\text{ms}$ in causal LM).

\paragraph{Operator results.}
At $H=8, D=64, L=262{,}144$, \emph{THOT Kernel (BF16 TC)} executes forward in \textbf{$34.71\,\text{ms}$} and forward+VJP in \textbf{$100.10\,\text{ms}$} with $1{,}132.1\,\text{MiB}$ static XLA compiler temporary allocation. Even in full FP32 precision, split accumulators and native grid batching cut forward+VJP latency from $480.83\,\text{ms}$ (\emph{Unfused Hub OT}) to $268.90\,\text{ms}$ while halving compiler temporary storage ($2{,}160.6 \to 1{,}132.1\,\text{MiB}$). Relative to \emph{Latent Softmax}, our BF16 Tensor-Core kernel is $2.22\times$ faster forward+VJP with $4.0\times$ smaller compiler temporaries ($1{,}132.1$ vs.\ $4{,}521.3\,\text{MiB}$; note that cuDNN FlashAttention requires slightly smaller static compiler temporaries at $1{,}040.0\,\text{MiB}$). At moderate sequence length ($B=2, L=4{,}096, H=8, D=32, r=32$, bidirectional), THOT beats cuDNN FlashAttention by $2.02\times$ forward ($0.544$ vs.\ $1.100\,\text{ms}$) and $3.27\times$ forward+VJP ($1.337$ vs.\ $4.368\,\text{ms}$). At the full 2-layer MoE model level ($D=32, L=262{,}144$, measured prior to split-factor kernel integration), THOT runs forward+grad in $139.26\,\text{ms}$ ($2{,}614.1\,\text{MiB}$ static temp), beating \emph{Latent Softmax + Sparsemax} ($144.34\,\text{ms}$, $2{,}851.1\,\text{MiB}$) and outpacing the dense FlashAttention MoE ($1{,}426.69\,\text{ms}$) by over $10\times$.

\begin{table}[t]
  \centering
  \caption{\textbf{Bidirectional operator and WikiText-4K attention-layer microbenchmarks on NVIDIA L4} (median of 5 synchronized runs; attention operator only, excluding QKV projections and gating/sorting). Memory reports XLA static compiler temporary buffer estimates, not runtime peak VRAM.}
  \label{tab:systems}
  \setlength{\tabcolsep}{5pt}
  \resizebox{\linewidth}{!}{%
  \begin{tabular}{@{}llrrr@{}}
    \toprule
    Geometry & Operator (Bidirectional) & Forward (ms) & Forward+VJP (ms) & Comp.\ Temp (MiB) \\
    \midrule
    $H=8, D=64, L=262{,}144$ & \textbf{THOT Kernel (BF16 TC)} & $\mathbf{34.708}$ & $\mathbf{100.097}$ & $1{,}132.1$ \\
    ($B=1, r=4, K=16$) & THOT Kernel (FP32) & $111.293$ & $268.897$ & $1{,}132.1$ \\
    & Unfused Hub OT & $109.278$ & $480.834$ & $2{,}160.6$ \\
    & Latent Softmax & $42.966$ & $221.889$ & $4{,}521.3$ \\
    & FlashAttention (cuDNN) & $2{,}227.004$ & $9{,}235.409$ & $\mathbf{1{,}040.0}$ \\
    \midrule
    $H=1, D=32, L=262{,}144$ & \textbf{THOT Kernel (BF16 TC)} & $\mathbf{1.684}$ & $\mathbf{5.380}$ & $74.3$ \\
    ($B=1, r=4, K=16$) & Unfused Hub OT & $7.175$ & $24.448$ & $142.0$ \\
    & Latent Softmax & $5.593$ & $22.964$ & $292.1$ \\
    & FlashAttention (cuDNN) & $279.524$ & $1{,}287.432$ & $\mathbf{66.0}$ \\
    \midrule
    WikiText-4K Layer & \textbf{THOT Kernel (BF16 TC)} & $\mathbf{0.544}$ & $\mathbf{1.337}$ & $33.0$ \\
    ($B=2, L=4{,}096, H=8$) & THOT Kernel (FP32) & $2.186$ & $5.138$ & $33.0$ \\
    ($D=32, r=32, K=16$) & Unfused Hub OT & $3.959$ & $12.795$ & $59.3$ \\
    & Latent Softmax & $1.045$ & $4.906$ & $80.8$ \\
    & FlashAttention (cuDNN) & $1.100$ & $4.368$ & $\mathbf{16.5}$ \\
    \bottomrule
  \end{tabular}}
\end{table}

\subsection{Strictly Causal Autoregressive Benchmarks}
\label{sec:exp-causal}

\begin{figure}[t]
  \centering
  \includegraphics[width=\linewidth]{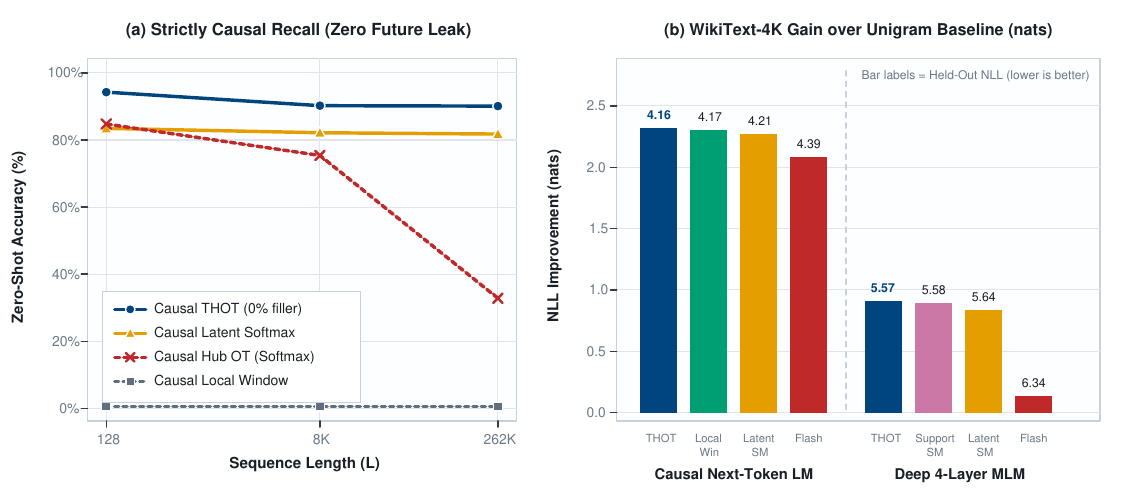}
  \caption{\textbf{Strictly causal associative recall and compact WikiText-103 4K language modeling.}
  \textbf{(a)} Under verified $0.0$ future leakage (annealed gate warmup over steps $1$--$256$), \emph{Causal THOT} sustains $90.10\%$ zero-shot recall at $L=262\text{K}$ ($0.00\%$ excess filler share), outperforming \emph{Causal Latent Softmax} ($81.77\%$) and \emph{Causal Hub OT (Softmax Gate)} ($32.81\%$).
  \textbf{(b)} On compact 4-layer Causal Next-Token LM ($512$ steps) and Deep 4-Layer MLM ($640$ steps) over WikiText-103 $4\text{K}$ documents, THOT achieves the lowest held-out NLL, with modest margins ($0.015$--$0.016$ nats) over the strongest nearby structured baselines.}
  \label{fig:causal-and-wikitext}
\end{figure}

Table~\ref{tab:causal} and Figure~\ref{fig:causal-and-wikitext}(a) summarize our autoregressive evaluations. On strictly causal associative recall (all $R=8$ bindings precede all $Q=8$ queries beyond the local window radius; 32 test sequences/seed, 768 queries across 3 seeds), all models train for $1{,}536$ steps at $L=128$ using an annealed auxiliary gate warmup ($\alpha$ linearly decayed from $1.0$ to $0.0$ over steps $1$--$256$, and $\alpha=0$ for steps $257$--$1{,}536$). \emph{Causal Local Window} fails ($0.52\%$) and softmax-gated causal OT drops to $32.81\%$ at $L=262\text{K}$ as distractors absorb $98.87\%$ of excess capacity. In contrast, \emph{Causal THOT} maintains $0.00\%$ excess filler share and sustains \textbf{$90.10\%$} accuracy at $L=262{,}144$ ($(91.41, 91.41, 87.50)\%$ across seeds; $0.564$ NLL), outperforming \emph{Causal Latent Softmax} ($81.77\%$, $1.162$ NLL) by $+8.3$ percentage points.

We also run a preliminary compact-scale causal LM comparison using 4-layer, 8-head models ($D=256$, $2{,}048$-token vocabulary, $r=32$, $K=16$, $C=32$, 4 MoE experts) on $L=4{,}096$ WikiText-103 documents (batch size $2$, 2 seeds, $512$ steps, unigram NLL $6.4770$). Held-out NLL is evaluated on intact $4\text{K}$ sequences (\emph{Intact}) and when documents are segmented into isolated $16 \times 256$-token blocks (\emph{Blocked}). \emph{Causal THOT} reaches the lowest intact NLL (\textbf{$4.1560$}: $4.1582 / 4.1538$ across seeds) and highest token accuracy (\textbf{$22.48\%$}), ahead of \emph{Causal Local Window} ($4.1706$: $4.1676 / 4.1736$) by a modest $0.0146$ nats, \emph{Causal Latent Softmax} ($4.2074$) by $0.0514$ nats, and \emph{Causal FlashAttention} ($4.3903$). Because \emph{Causal Local Window} ($r=32$) itself gains $+0.0080$ nats on intact over blocked documents due to artificial 256-token boundary truncation, the intact-minus-blocked difference (\emph{Ctx Gain}) mixes boundary effects with genuine long-range context use and should be interpreted cautiously. On this unfused multi-layer training path, \emph{Causal THOT} takes $211.41\,\text{ms/step}$ ($1.67\times$ slower than \emph{Causal FlashAttention} at $126.45\,\text{ms/step}$).

\begin{table}[t]
  \centering
  \caption{\textbf{Strictly causal associative recall ($L=128\to 262\text{K}$, 3 seeds, 768 queries; gate supervision $\alpha$ annealed $1.0\to 0.0$ over steps $1$--$256$) and preliminary causal WikiText-4K LM (4-layer $D=256$, $V=2{,}048$, 2 seeds $\times 512$ steps).} Every model verifies $0.0$ future-token leakage; full-model step times reflect the unfused reference path.}
  \label{tab:causal}
  \setlength{\tabcolsep}{4pt}
  \resizebox{\linewidth}{!}{%
  \begin{tabular}{@{}lccccccc@{}}
    \toprule
    & \multicolumn{2}{c}{Causal Recall ($\alpha: 1\!\to\!0$)} & \multicolumn{5}{c}{Causal WikiText-4K LM (Unigram $6.4770$)} \\
    \cmidrule(lr){2-3} \cmidrule(lr){4-8}
    Causal Model & $128 \to 8\text{K} \to 262\text{K}$ & $262\text{K}$ Ex.\ Fill & Step (ms) & Intact & Blocked & Ctx Gain & Acc \\
    \midrule
    \textbf{Causal THOT} & $\mathbf{94.27 / 90.23 / 90.10\%}$ & $\mathbf{0.00\%}$ & $211.41$ & $\mathbf{4.1560}$ & $\mathbf{4.1724}$ & $+0.0165$ & $\mathbf{22.48\%}$ \\
    Causal Latent Softmax & $83.46 / 82.16 / 81.77\%$ & $0.00\%$ & $151.03$ & $4.2074$ & $4.2330$ & $\mathbf{+0.0256}$ & $21.57\%$ \\
    Causal Local Window & $0.52 / 0.52 / 0.52\%$ & $99.99\%$ & $\mathbf{121.54}$ & $4.1706$ & $4.1786$ & $+0.0080$ & $22.33\%$ \\
    Causal FlashAttention & --- & --- & $126.45$ & $4.3903$ & $4.4137$ & $+0.0234$ & $18.21\%$ \\
    Causal Hub OT (Softmax Gate) & $84.77 / 75.39 / 32.81\%$ & $98.87\%$ & --- & --- & --- & --- & --- \\
    \bottomrule
  \end{tabular}}
\end{table}

\subsection{Natural-Language Evaluation: Multi-Needle Prose Retrieval and WikiText-4K MLM}
\label{sec:exp-wikitext}

To assess behavior on natural English text beyond synthetic token streams, we evaluate two compact benchmarks derived from WikiText-103 \citep{merity2017pointer}:

\paragraph{Zero-shot multi-needle retrieval in WikiText-103 prose (\texorpdfstring{$\alpha=0$}{alpha=0}).}
We embed $R=8$ constructed key-value pairs into contiguous WikiText-103 passages using three zero-occurrence real-vocabulary punctuation markers ($\langle\text{BIND}\rangle$, $\langle\text{VAL}\rangle$, $\langle\text{QRY}\rangle$). Each key is a 2-token pair $(k_1, k_2)$ sampled without replacement from $16 \times 32 = 512$ real-vocabulary tokens, and each value $v$ is drawn from $256$ real-vocabulary tokens that also occur naturally in the surrounding prose. Each binding appears as the 5-token span $[\langle\text{BIND}\rangle, k_1, k_2, \langle\text{VAL}\rangle, v]$ inside the source arc $[L/10, 3L/8]$, and each corresponding query appears as $[\langle\text{QRY}\rangle, k_1, k_2, \langle\text{VAL}\rangle, \langle\text{MASK}\rangle]$ inside the disjoint query arc $[5L/8, 9L/10]$ (strictly beyond the multi-layer local window radius). Cross-entropy loss and accuracy are evaluated strictly on the $\langle\text{MASK}\rangle$ positions over the full $2{,}048$-token vocabulary without any gate supervision ($\alpha=0$). Models train for $512$ steps at $L=512$ across two seeds ($1711, 2917$) and are evaluated zero-shot on 16 passages/seed (256 total queries) at $L \in \{512, 8{,}192\}$ and 8 passages/seed (128 total queries) at $L=65{,}536$.

Table~\ref{tab:wikitext}(left) and Figure~\ref{fig:dilution-and-retrieval}(b) report the results. The short-trained ($L=512$) RoPE \emph{FlashAttention (cuDNN)} baseline reaches $99.22\%$ in-distribution at $L=512$, then drops to $0.78\%$ ($8.0463$--$8.1803$ NLL) at $8\text{K}$ and $64\text{K}$. Because this dense baseline uses rotary position embeddings (RoPE; \citealp{su2024roformer}) evaluated up to $128\times$ beyond its training length, its collapse reflects both positional extrapolation failure and attention dispersion rather than normalization dilution in isolation (the controlled isolation of gating dilution remains Table~\ref{tab:synth-main}). Among the sparse-gated linear-time architectures on the identical task, legacy $4+2$ OT ($83.59\%$), \emph{Latent Softmax + Sparsemax} ($79.69\%$), and \emph{Support Softmax + Sparsemax} ($78.91\%$) preserve functional retrieval at $64\text{K}$, whereas \textbf{THOT (1+2, Sparsemax)} reaches \textbf{$100.00\%$} (64/64 on both seeds; 128/128 queries, $0.0514$ NLL) at $L=65{,}536$.

\paragraph{Preliminary 4-layer WikiText-4K masked language modeling.}
We next evaluate bidirectional 4-layer ($D=256, H=8, V=2{,}048, r=32, K=16$) models on 256 held-out $4{,}096$-token WikiText-103 confirmation documents after $640$ training steps ($\sim 15\%$ masked across 1--3 word geometric spans, learning rate $10^{-3}$, 2 seeds, unigram NLL $6.4788$). Four uniform stabilization practices are applied across all models (Appendix~\ref{app:details}): non-zero learned mask embeddings, unit output logit scaling, independent query/key initializations with full rotary coverage ($d_{\text{rope}}=32$, base $10{,}000$) for dense FlashAttention, and short 1--3 word spans. Under this compact setup (Table~\ref{tab:wikitext}, right), \textbf{THOT (1+2, Sparsemax)} obtains the lowest intact NLL (\textbf{$5.5657$}: $5.5678 / 5.5636$ across seeds, $10.07\%$ accuracy), modestly ahead of \emph{Support Softmax + Sparsemax} ($5.5819$: $5.5623 / 5.6016$, a $0.0162$-nat mean difference with overlapping single-seed values) and \emph{Latent Softmax + Sparsemax} ($5.6430$). Full-RoPE \emph{FlashAttention} reaches only $6.3424$ NLL (barely below the $6.4788$ unigram baseline) within $640$ updates at batch size $2$; because that gap could reflect undertraining or hyperparameter sensitivity of dense attention at small step budgets, we treat these MLM numbers as preliminary diagnostics rather than evidence of broad language-modeling superiority. On this unfused path, THOT runs at $174.15\,\text{ms/step}$ ($1.28\times$ slower than dense FlashAttention at $136.32\,\text{ms/step}$).

\begin{table}[t]
  \centering
  \caption{\textbf{Natural-language evaluation on WikiText-103} (2 seeds). \emph{Left:} Zero-shot multi-needle prose retrieval ($R=8, \alpha=0$, trained at $L=512$; 256 queries at $512$/$8\text{K}$, 128 queries at $64\text{K}$). \emph{Right:} Preliminary 4-layer $D=256$ WikiText-4K MLM (256 held-out docs, $640$ steps, Unigram NLL $6.4788$, unfused step times).}
  \label{tab:wikitext}
  \setlength{\tabcolsep}{4pt}
  \resizebox{\linewidth}{!}{%
  \begin{tabular}{@{}lcc|ccccc@{}}
    \toprule
    & \multicolumn{2}{c|}{Multi-Needle Prose ($\alpha=0$)} & \multicolumn{5}{c}{Deep WikiText-4K MLM (Unigram $6.4788$)} \\
    Method & $512 \to 8\text{K} \to 64\text{K}$ & $64\text{K}$ NLL & Step (ms) & Intact & Blocked & Ctx Gain & Acc \\
    \midrule
    \textbf{THOT (1+2, Sparsemax)} & $\mathbf{95.31 / 98.05 / 100.0\%}$ & $\mathbf{0.0514}$ & $174.15$ & $\mathbf{5.5657}$ & $\mathbf{5.5808}$ & $+0.0151$ & $\mathbf{10.07\%}$ \\
    Hub OT (4+2, Sparsemax) & $84.77 / 81.64 / 83.59\%$ & $0.5269$ & --- & --- & --- & --- & --- \\
    Support Softmax + Sparsemax & $81.25 / 79.30 / 78.91\%$ & $0.7507$ & $139.45$ & $5.5819$ & $5.7954$ & $\mathbf{+0.2135}$ & $9.98\%$ \\
    Latent Softmax + Sparsemax & $88.28 / 87.50 / 79.69\%$ & $0.8376$ & $136.78$ & $5.6430$ & $5.6663$ & $+0.0233$ & $9.41\%$ \\
    FlashAttention (cuDNN, RoPE) & $99.22 / 0.78 / 0.78\%$ & $8.1803$ & $\mathbf{136.32}$ & $6.3424$ & $6.4496$ & $+0.1072$ & $4.07\%$ \\
    \bottomrule
  \end{tabular}}
\end{table}

\section{Related Work}
\label{sec:related}

\paragraph{Sparse, latent, slot-memory, and linear long-context architectures.}
Longformer \citep{beltagy2020longformer} and BigBird \citep{zaheer2020bigbird} introduced combined local sliding windows and global tokens for linear-time Transformers, while Perceiver \citep{jaegle2021perceiver} routes inputs through learned latent bottleneck arrays and Vision Transformers employ global register tokens as routing scratchpads \citep{darcet2024vision}. Complementary linear-time methods---including linear attention \citep{katharopoulos2020transformers}, selective state-space models \citep{gu2024mamba}, gated linear attention \citep{yang2024gla}, and Gated Slot Attention (GSA) \citep{zhang2024gsa}---compress history into recurrent or slot-based memories via parallel scans; GSA is particularly relevant to our causal extension (\S\ref{sec:causal}), as it targets associative recall via compact recurrent slots with context-dependent reading, whereas Causal THOT balances chunk writes and prefix reads through semi-relaxed optimal transport coupled with explicit local windows and thresholded token capacities. \citet{arora2024zoology} highlighted recall deficits in sub-quadratic models, while long-context evaluations document U-shaped positional drop-offs \citep{liu2024lost}, multi-needle degradation \citep{hsieh2024ruler}, and sensitivity to positional extrapolation or attention sinks \citep{press2022train,peng2024yarn,xiao2024efficient}.

\paragraph{Differentiable optimal transport and low-rank Sinkhorn attention.}
Entropic regularization \citep{cuturi2013sinkhorn,peyre2019computational} made Sinkhorn scaling \citep{sinkhornknopp1967,knight2008sinkhorn} practical inside neural layers, motivating feature matching \citep{sarlin2020superglue}, unbalanced/semi-relaxed transport \citep{chizat2018scaling}, smooth quadratic transport \citep{blondel2018smooth}, doubly-stochastic attention \citep{tay2020sparse,sander2022sinkformers}, warm-start initialization \citep{thornton2023initialization}, and unrolled or implicit differentiation \citep{luise2018differential,eisenberger2022unified}. Closely related to our hub-mediated architecture is \textbf{LOTFormer} \citep{shahbazi2025lotformer}, which constructs query-to-pivot and pivot-to-key optimal transport problems through a small learned intermediate measure and composes them into an $O(nr)$ low-rank doubly-stochastic attention operator. THOT differs from LOTFormer in five specific ways: (i) coupling global hubs jointly with a local sliding-window band inside a single sparse graph $\Omega$, (ii) replacing uniform marginals with input-dependent bounded-support token capacities $(a,b)$ to eliminate distractor dilution, (iii) introducing the $1+2$ annealed row-closed finite-sweep schedule (and prior-preserving semi-relaxed variant), (iv) deriving an exact rank-$(D+2R+1)$ One-Reference-Tile adjoint implemented in split-factor BF16 Tensor-Core Triton kernels, and (v) providing a strictly causal chunked exclusive-prefix construction. Most directly, this paper builds on \citet{forde2026blockwise} (block-wise Sinkhorn tail refinement) and \citet{forde2026topology}
{} (fixed-support gradient mixing).

\paragraph{Sparse attention normalization and anti-dilution.}
Sparsemax \citep{martins2016softmax} and $\alpha$-entmax \citep{correia2019adaptively} replace softmax with compact-support simplex projections, while \citet{wortsman2023replacing} studied point-wise $\ReLU$ attention scaling. Most closely aligned with our anti-dilution motivation are \citet{velickovic2025softmax}, who proved that standard softmax must disperse on sharp size generalization and studied adaptive temperature, and \textbf{Long-Context Generalization with Sparse Attention} \citep{vasylenko2025longcontext}, who showed that distractors accumulate softmax mass and used exact-zero sparse attention normalization to achieve up to $\sim 1{,}000\times$ synthetic length extrapolation and improved language-model generalization. Whereas \citet{vasylenko2025longcontext} apply exact sparsity directly to the pairwise attention weights, THOT applies exact-zero sparsity at the \emph{OT marginal capacity gate} (via $\sparsemax$ or $O(L)$ thresholded $\ReLU$), pairing sparse capacity allocation with a linear-time hub-and-local transport graph and custom Triton kernels.

\paragraph{IO-aware attention and OT kernels.}
The FlashAttention family \citep{dao2022flashattention,dao2023flashattention2,shah2024flashattention3} avoids $O(L^2)$ HBM materialization via SRAM tiling and mixed-precision Tensor Cores implemented in CUDA or Triton \citep{triton2019tillet}. For dense entropic optimal transport, FlashSinkhorn \citep{flashsinkhorn2026} recently introduced IO-aware streaming Triton kernels for full $N\times N$ Sinkhorn scaling. Our split-factor Triton kernels apply analogous SRAM tiling and BF16 Tensor Cores directly to the \emph{sparse hub-and-spoke graph} while keeping dual potentials strictly FP32 and compressing all finite-tail score gradients into two width-$D$ reference-tile passes.

\section{Limitations, Reproducibility, and Conclusion}
\label{sec:conclusion}

\paragraph{Limitations.}
We highlight five boundaries of the current study:
\begin{enumerate}[leftmargin=1.5em]
  \item \textbf{Operator vs.\ full-model kernel integration:} Our split-factor BF16 Tensor-Core kernels were benchmarked on standalone bidirectional operators ($100.10\,\text{ms}$ at $262\text{K}$; $1.337\,\text{ms}$ vs.\ $4.368\,\text{ms}$ cuDNN FlashAttention at $4\text{K}$), whereas the multi-layer WikiText-4K training loops ran on our unfused path ($174.15\,\text{ms/step}$ MLM vs.\ $136.32\,\text{ms}$ dense; $211.41\,\text{ms/step}$ causal vs.\ $126.45\,\text{ms}$ dense). Faster end-to-end language-model training has therefore not yet been demonstrated and requires wiring the Triton kernels into full multi-layer bidirectional and causal stacks.
  \item \textbf{Static compiler vs.\ runtime memory:} Reported memory figures reflect static XLA compiler temporary estimates rather than dynamic peak runtime VRAM profiles, and the $4.0\times$ compiler-temporary reduction is strictly relative to \emph{Latent Softmax} (not FlashAttention).
  \item \textbf{Fixed-binding retrieval vs.\ information capacity scaling:} Our $262\text{K}$ extrapolation benchmarks hold $R=8$ relevant bindings fixed while expanding distractors; under hub overload ($R=32$ vs.\ $K=16$), recall falls to $38\%$--$40\%$. Scaling $R$, $K$, and $L$ jointly remains important future work.
  \item \textbf{Preliminary LM scale and dense baseline diagnostics:} Our WikiText-4K experiments use compact 4-layer ($D=256$, $V=2{,}048$, $512$--$640$ steps, 2 seeds) models with modest $0.015$--$0.016$ nat margins over nearby structured baselines, and our prose-needle dense baseline conflates RoPE extrapolation with attention dispersion. Diagnosing why the dense MLM baseline remained near unigram ($6.3424$ vs.\ $6.4788$) and adding content-only (no-RoPE) dense retrieval controls are priorities for a follow-up revision.
  \item \textbf{Variant trade-offs:} Sorting-free $O(L)$ thresholded $\ReLU$ relies on $\alpha=0.05$ to match $\sparsemax$ at $262\text{K}$ ($98.83\%$ vs.\ $80.60\%$ at $\alpha=0$), Causal Recall uses an annealed gate warmup over steps $1$--$256$, and non-annealed row-closed bipartite OT outperforms annealed THOT on nonzero collision stress under $\alpha=0$.
\end{enumerate}

\paragraph{Reproducibility and Code Availability.}
All experiments ran on NVIDIA L4 GPUs in JAX (0.11.1, jaxlib 0.11.1, NumPy 2.4.6) with fixed random seeds, deterministic data generators, and matched parameter budgets across compared arms. Key experimental configurations, evaluation counts, and individual per-seed results appear in Appendix~\ref{app:details}. A concise standalone JAX/Triton reference implementation covering the capacity gates, batched split-factor THOT kernels (with an exact pure-JAX reference for CPU verification), and strictly causal chunked operator accompanies this manuscript. Notice on \texttt{arXiv:2609.07954}: this manuscript extends our earlier fixed-support transport analysis with sparse capacity gating, finite-sweep annealing, exact Triton adjoint kernels, causal chunking, and empirical long-context benchmarks.

\paragraph{Conclusion.}
Topology Hub Optimal Transport pairs bounded-support capacity gating, a $1+2$ annealed row-closed Sinkhorn schedule, an exact rank-$(D+2R+1)$ One-Reference-Tile adjoint implemented in split-factor BF16 Tensor-Core Triton kernels, and a strictly causal exclusive-prefix chunked variant. Controlled ablations isolate sparse capacity allocation as the decisive factor preventing $1/(1+\alpha L)$ distractor collapse ($3.52\% \to 97.40\%$ pure cross-entropy recall at $L=262\text{K}$), while operator microbenchmarks show linear-time bidirectional OT surpassing cuDNN FlashAttention already at $L=4{,}096$ ($1.34$ vs.\ $4.37\,\text{ms}$) and running $2.2\times$ faster than latent softmax at $L=262\text{K}$. Integrating these kernels into full multi-layer causal and bidirectional language models at larger scale is the natural next step.

\appendix
\section{Exact Adjoint Derivation and Experimental Details}
\label{app:details}

\subsection{Exact One-Reference-Tile Adjoint for Algorithm 1}
\label{app:adjoint}

We derive the exact reverse-mode vector-Jacobian product (VJP) implemented by our batched Triton kernel for the bidirectional $1+2$ annealed row-closed operator in \hyperlink{alg:thot}{Algorithm~1} ($R = 2$ differentiated bipartite tail sweeps with $B_{HH} = -20$, following a detached aperiodic warm-start potential $v_1$). Let $A = \{i : a_i > 0\}$ and $B = \{j : b_j > 0\}$ denote active rows and columns, with $|A| = \sum_{i} \mathbf{1}[i \in A]$.

\paragraph{Forward states cached for backward.}
In Algorithm~1, Sweeps $t \in \{2, 3\}$ iterate from $v_1 = \mathrm{detach}(v_1)$ under the bipartite edge scores $S$:
\begin{align}
  u^{(t)}_{\text{raw},i} &= \ln a_i - \operatorname{LSE}_{j \in \Omega(i)}(S_{ij} + v_{t-1,j}), &
  s_t &= \frac{1}{|A|} \sum_{i \in A} u^{(t)}_{\text{raw},i}, \nonumber\\
  u_{t,i} &= u^{(t)}_{\text{raw},i} - s_t, &
  v_{t,j} &= \ln b_j - \operatorname{LSE}_{i \in \Omega^\top(j)}(S_{ij} + u^{(t)}_{\text{raw},i}) + s_t,
  \label{eq:app-forward-sweep}
\end{align}
followed by exact row closure $u_{\text{out},i} = \ln a_i - \operatorname{LSE}_{j \in \Omega(i)}(S_{ij} + v_{3,j})$, terminal plan $P^{\text{out}}_{ij} = \exp(S_{ij} + u_{\text{out},i} + v_{3,j})$, unnormalized aggregate $Z = P^{\text{out}} V$, and row-stochastic attention output $Y_i = Z_i / a_i$. Forward execution caches $(Z, v_1, u^{(2)}_{\text{raw}}, s_2, v_2, u^{(3)}_{\text{raw}}, s_3, v_3, u_{\text{out}})$ (total storage $O(N)$ scalars per head plus $Z$), eliminating all five recomputation log-sum-exp passes during backward.

\paragraph{Step 1: Quotient normalization and terminal value pass.}
Given upstream loss gradient $\bar{Y} = \partial \ell / \partial Y$, differentiating $Y_i = Z_i / a_i$ yields the unnormalized aggregate cotangent $\bar{Z}_i = \bar{Y}_i / a_i$ together with the direct normalization marginal contribution $\bar{a}^{\text{norm}}_i = -(\bar{Y}_i^\top Y_i) / a_i$. Differentiating $Z = P^{\text{out}} V$ gives the value gradient via one width-$D$ transposed sparse pass:
\begin{equation}
  \nabla_V = (P^{\text{out}})^\top \bar{Z} = W^\top \bar{Y},
\end{equation}
and nonzero terminal log-potential sensitivities
\begin{equation}
  g_{u,i} = \bar{Z}_i^\top Z_i = \bar{Y}_i^\top Y_i, \qquad
  g_{v,j} = V_j^\top (\nabla_V)_j,
  \label{eq:app-terminal-gu-gv}
\end{equation}
with direct score gradient $E^{\text{direct}}_{ij} = P^{\text{out}}_{ij}\, (\bar{Z}_i^\top V_j)$.

\paragraph{Step 2: Terminal row-closure reverse step.}
Differentiating $u_{\text{out},i} = \ln a_i - \operatorname{LSE}_j(S_{ij} + v_{3,j})$ using $\sum_j P^{\text{out}}_{ij} = a_i$ yields
\begin{equation}
  r_{\text{out},i} = \frac{g_{u,i}}{a_i} \quad (i \in A), \qquad
  \bar{a} \leftarrow \bar{a}^{\text{norm}} + r_{\text{out}}, \qquad
  g_v \leftarrow g_v - (P^{\text{out}})^\top r_{\text{out}},
  \label{eq:app-row-closure}
\end{equation}
and adds the rank-1 score contribution $-P^{\text{out}}_{ij}\, r_{\text{out},i}$ (note that $\bar{a}^{\text{norm}}_i + r_{\text{out},i} = 0$, confirming that row-stochastic $Y_i = Z_i/a_i$ has zero net direct sensitivity to $a_i$ outside the tail sweeps).

\paragraph{Step 3: Exact reverse recurrence through Sweeps $t = 3, 2$.}
Define the intermediate plans $P^{(t)}_{ij} = \exp(S_{ij} + u_{t,i} + v_{t,j})$ (whose column sums satisfy $(P^{(t)})^\top \one = b$) and $P^{(t)}_{\text{raw},ij} = \exp(S_{ij} + u^{(t)}_{\text{raw},i} + v_{t-1,j})$ (whose row sums satisfy $P^{(t)}_{\text{raw}} \one = a$). For $t = 3$ and then $t = 2$ (noting $g_u = 0$ on entry since $u_3, u_2$ are unused outside their own sweep):
\begin{align}
  c_{t,j} &= \frac{g_{v,j}}{b_j} \;\; (j \in B), &
  d_{c,t} &= \frac{1}{|A|} \sum_{j \in B} g_{v,j}, &
  r_{t,i} &= \frac{d_{c,t} - (P^{(t)} c_t)_i}{a_i} \;\; (i \in A), \label{eq:app-rev-step}\\
  \bar{a} &\leftarrow \bar{a} + r_t, &
  \bar{b} &\leftarrow \bar{b} + c_t, &
  g_v &\leftarrow -(P^{(t)}_{\text{raw}})^\top r_t \;\; (\text{for } t = 3 \text{ only}), \nonumber
\end{align}
where $v_1$ is detached so no propagation past $t=2$ is required. Each sweep $t$ contributes $-P^{(t)}_{ij} c_{t,j} - P^{(t)}_{\text{raw},ij} r_{t,i}$ to the edge score adjoint $\bar{S}_{ij}$. Across Steps 2--3, the entire vector recurrence requires only \textbf{four} lightweight 1D scalar sparse matvecs: $(P^{\text{out}})^\top r_{\text{out}}$, $P^{(3)} c_3$, $(P^{(3)}_{\text{raw}})^\top r_3$, and $P^{(2)} c_2$.

\paragraph{Step 4: Exact diagonal-rescaling factorization onto $P^{\text{out}}$.}
Crucially, because intermediate plans $P^{(t)}$ and $P^{(t)}_{\text{raw}}$ differ from the terminal plan $P^{\text{out}}$ on $\Omega$, their score contributions do \emph{not} collapse to a rank-2 additive row-plus-column term $\alpha_i + \beta_j$. Instead, every intermediate plan is an exact rank-1 diagonal rescaling of $P^{\text{out}}$:
\begin{align}
  P^{(t)}_{ij} &= P^{\text{out}}_{ij}\, \underbrace{\exp\!\big((u_{t,i} - u_{\text{out},i}) - \mu_{u,t}\big)}_{e^u_{t,i}} \, \underbrace{\exp\!\big((v_{t,j} - v_{3,j}) + \mu_{u,t}\big)}_{e^v_{t,j}}, \nonumber\\
  P^{(t)}_{\text{raw},ij} &= P^{\text{out}}_{ij}\, \underbrace{\exp\!\big((u^{(t)}_{\text{raw},i} - u_{\text{out},i}) - \mu_{\text{raw},t}\big)}_{e^{\text{raw}}_{t,i}} \, \underbrace{\exp\!\big((v_{t-1,j} - v_{3,j}) + \mu_{\text{raw},t}\big)}_{e^{\text{prev}}_{t,j}},
  \label{eq:app-orbit-rescale}
\end{align}
where mean potential shifts $\mu_{u,t} = \frac{1}{|A|}\sum_{i\in A}(u_{t,i} - u_{\text{out},i})$ and $\mu_{\text{raw},t} = \frac{1}{|A|}\sum_{i\in A}(u^{(t)}_{\text{raw},i} - u_{\text{out},i})$ cancel identically inside the product while preventing floating-point overflow. Collecting the terminal row-closure term and the $2R = 4$ rescaled tail terms into two $(N \times 5)$ factor matrices
\begin{align}
  L^{(s)} &= \big[\, -r_{\text{out}},\;\; -e^u_3,\;\; -e^{\text{raw}}_3 \odot r_3,\;\; -e^u_2,\;\; -e^{\text{raw}}_2 \odot r_2 \,\big] \in \mathbb{R}^{N \times (2R+1)}, \nonumber\\
  R^{(s)} &= \big[\, \one_B,\;\; e^v_3 \odot c_3,\;\; e^{\text{prev}}_3,\;\; e^v_2 \odot c_2,\;\; e^{\text{prev}}_2 \,\big] \in \mathbb{R}^{N \times (2R+1)},
  \label{eq:app-ls-rs}
\end{align}
the \emph{exact} total edge score adjoint across row closure and Sweeps 2--3 is
\begin{equation}
  \bar{S}_{ij} = P^{\text{out}}_{ij} \left( \bar{Z}_i^\top V_j + \sum_{m=1}^{2R+1} L^{(s)}_{i,m}\, R^{(s)}_{j,m} \right).
  \label{eq:app-exact-sbar}
\end{equation}
Hence $\nabla_Q = s_{\text{scale}}\, \bar{S} K$ (together with fused node-type bias sums $\nabla_{B_{c,c'}} = \sum_{(i,j)\in\Omega_{c,c'}} \bar{S}_{ij}$, clamping $\nabla_{B_{HH}} = 0$) and $\nabla_K = s_{\text{scale}}\, \bar{S}^\top Q$ execute in \textbf{two} width-$D$ split-factor sparse passes over the single reference plan $P^{\text{out}}$ (down from $12$ width-$D$ score-gradient passes in naive unrolled differentiation). In our GPU verification on $L \in \{128, 256\}$ ($H \in \{2,4\}, D \in \{32,64\}$), this exact custom VJP matches the unfused reference implementation to $\le 4.47\times 10^{-8}$ max absolute error ($\le 5.02\times 10^{-7}$ relative $L_2$) in FP32 and $\le 3.27\times 10^{-4}$ in BF16 Tensor-Core mode, and matches pure-JAX unrolled dense autodiff within $\le 6.56\times 10^{-7}$ in FP32 across all six input gradients $(\nabla_Q, \nabla_K, \nabla_V, \nabla_a, \nabla_b, \nabla_B)$.

\subsection{Experimental Configurations, Per-Seed Results, and Stabilization Notes}

\begin{table}[htbp]
  \centering
  \small
  \caption{\textbf{Key experimental configurations, optimizer schedules, and evaluation sample counts.}}
  \label{tab:hparams}
  \begin{tabular}{@{}l p{0.71\linewidth}@{}}
    \toprule
    Benchmark & Configuration, Schedule, and Evaluation Details \\
    \midrule
    Hardware / Software & NVIDIA L4 GPU; JAX 0.11.1, jaxlib 0.11.1, NumPy 2.4.6; microbenchmarks report median of 5 device-synchronized repeats after compilation warmup \\
    Synthetic Recall & 2-layer MoE ($D=32$, 4 experts, hidden $48$), $K=16, r=4, M=8.0$, $R=Q=8$ bindings/queries; AdamW, $1{,}536$ steps at $L=128$, 3 seeds ($1711, 2917, 4129$); evaluated on 32 held-out sequences/seed (256 queries/seed, 768 total) \\
    Kernel Microbench & Standalone bidirectional attention operators; BF16 $(Q,K,V)$ inputs; THOT BF16 TC (BF16 matmul, FP32 acc.\ \& potentials), THOT FP32, Unfused Hub OT (FP32), Latent Softmax (FP32), cuDNN FlashAttention (BF16); excludes QKV projections and gating/sorting \\
    Causal Recall & Same 2-layer MoE backbone ($D=32, K=16, r=4, C=32$), $1{,}536$ steps at $L=128$, cosine LR (peak $2\times 10^{-3}$), gate supervision $\alpha$ linearly annealed $1.0 \to 0.0$ over steps $1$--$256$; 3 seeds, 32 sequences/seed (768 queries) \\
    Causal WT-4K LM & 4-layer MoE ($H=8, D=256, V=2{,}048, r=32, K=16, C=32$), AdamW ($\text{lr}=2.5\times 10^{-3}$, $\beta_2=0.98$, weight decay $0.01$, cosine schedule), $B=2$, $512$ steps, 2 seeds ($1711, 2917$), evaluated on held-out confirmation split \\
    WikiText Prose Needles & 4-layer ($H=8, D=256, V=2{,}048, r=32, K=16$), $R=8$ needles, $\alpha=0$, $512$ steps at $L=512$ ($\text{lr}=3\times 10^{-3}$), 2 seeds ($1711, 2917$); evaluated on 16 passages/seed (256 queries) at $512$/$8\text{K}$ and 8 passages/seed (128 queries) at $64\text{K}$ \\
    WikiText-4K MLM & 4-layer ($H=8, D=256, V=2{,}048, r=32, K=16$), AdamW ($\text{lr}=10^{-3}, \beta_2=0.95$), $B=2$, $640$ steps, $\sim 15\%$ 1--3 word span masking, 2 seeds ($1711, 2917$), evaluated on 256 held-out $4{,}096$-token documents \\
    \bottomrule
  \end{tabular}
\end{table}

\begin{table}[htbp]
  \centering
  \small
  \caption{\textbf{Individual per-seed outcomes across primary benchmarks} (seeds 1711 / 2917 / 4129 for 3-seed synthetic and causal recall; seeds 1711 / 2917 for 2-seed WikiText-103 evaluations).}
  \label{tab:per-seed}
  \resizebox{\linewidth}{!}{%
  \begin{tabular}{@{}llcc@{}}
    \toprule
    Benchmark & Model / Variant & Per-Seed Values & Mean \\
    \midrule
    Synthetic Recall $262\text{K}$ ($\alpha=0.05$) & THOT (1+2, Sparsemax) & $98.83 / 98.44 / 99.22\%$ & $98.83\%$ \\
    & Bipartite OT (3-sweep, RC) & $98.44 / 97.27 / 97.66\%$ & $97.79\%$ \\
    & THOT (1+2, ReLU-Thresh, $O(L)$) & $97.66 / 99.61 / 99.22\%$ & $98.83\%$ \\
    & THOT-SR ($\gamma_b=0.92$ / $0.88$) & $96.48 / 98.44 / 97.27\%$ \;|\; $99.22 / 97.66 / 95.70\%$ & $97.40\%$ / $97.53\%$ \\
    \midrule
    Synthetic Recall $262\text{K}$ ($\alpha=0.00$) & THOT (1+2, Sparsemax) & $96.88 / 97.27 / 98.05\%$ & $97.40\%$ \\
    & Bipartite OT (3-sweep, RC) & $99.61 / 99.22 / 89.06\%$ & $95.96\%$ \\
    & THOT (1+2, ReLU-Thresh, $O(L)$) & $81.25 / 80.08 / 80.47\%$ & $80.60\%$ \\
    & THOT-SR ($\gamma_b=0.92$ / $0.88$) & $98.83 / 99.61 / 88.28\%$ \;|\; $81.64 / 98.05 / 85.94\%$ & $95.57\%$ / $88.54\%$ \\
    \midrule
    Causal Recall $262\text{K}$ ($\alpha: 1\!\to\!0$) & Causal THOT & $91.41 / 91.41 / 87.50\%$ & $90.10\%$ \\
    & Causal Latent Softmax & $82.81 / 76.95 / 85.55\%$ & $81.77\%$ \\
    & Causal Hub OT (Softmax Gate) & $33.98 / 33.98 / 30.47\%$ & $32.81\%$ \\
    \midrule
    WikiText Multi-Needle $64\text{K}$ ($\alpha=0$) & THOT (1+2, Sparsemax) & $100.00 / 100.00\%$ (64/64, 64/64) & $100.00\%$ \\
    & Hub OT (4+2, Sparsemax) & $76.56 / 90.62\%$ & $83.59\%$ \\
    & Latent Softmax + Sparsemax & $81.25 / 78.12\%$ & $79.69\%$ \\
    & Support Softmax + Sparsemax & $75.00 / 82.81\%$ & $78.91\%$ \\
    & FlashAttention (cuDNN, RoPE) & $1.56 / 0.00\%$ & $0.78\%$ \\
    \midrule
    Causal WikiText-4K LM (Intact NLL) & Causal THOT & $4.1582 / 4.1538$ & $4.1560$ \\
    & Causal Local Window & $4.1676 / 4.1736$ & $4.1706$ \\
    & Causal Latent Softmax & $4.2122 / 4.2027$ & $4.2074$ \\
    & Causal FlashAttention & $4.4043 / 4.3763$ & $4.3903$ \\
    \midrule
    Deep WikiText-4K MLM (Intact NLL) & THOT (1+2, Sparsemax) & $5.5678 / 5.5636$ & $5.5657$ \\
    & Support Softmax + Sparsemax & $5.5623 / 5.6016$ & $5.5819$ \\
    & Latent Softmax + Sparsemax & $5.6093 / 5.6768$ & $5.6430$ \\
    & FlashAttention (cuDNN, RoPE) & $6.3156 / 6.3691$ & $6.3424$ \\
    \bottomrule
  \end{tabular}}
\end{table}

\paragraph{WikiText-4K optimization stabilization details.}
In preliminary MLM prototypes, four initialization interactions selectively degraded baselines: (i) zero-initialized masked token embeddings produced zero vectors entering RMSNorm ($\approx 10^3\times$ backward Jacobian amplification), triggering global gradient-norm clipping; (ii) applying $1/\sqrt{D}$ scaling after unit-norm embeddings doubly attenuated output logits by $16\times$; (iii) nearly aligned query and key initializations combined with short-period partial RoPE locked dense FlashAttention onto the diagonal self-token; and (iv) overly long 10-word corruption spans hindered early optimization. Applying non-zero learned mask embeddings, unit logit scaling, independent query/key projections with full rotary coverage ($d_{\text{rope}}=32$, base $10{,}000$), and 1--3 word geometric masking spans uniformly addresses all four issues across every baseline and THOT model.

{\footnotesize
\setlength{\bibsep}{2pt plus 0.5pt}
\bibliographystyle{plainnat}
\bibliography{references}

@article{forde2026blockwise,
  title={Block-Wise Differentiable {Sinkhorn} Attention: Tail-Refinement Gradients with a Gap-Aware Dustbin Bridge},
  author={Forde, Dylan B.},
  journal={arXiv preprint arXiv:2605.08123},
  year={2026},
  doi={10.48550/arXiv.2605.08123}
}

@article{forde2026topology,
  title={Support Topology and Gradient Mixing in {Sinkhorn} Layers},
  author={Forde, Dylan B.},
  journal={arXiv preprint arXiv:2609.07954},
  year={2026},
  doi={10.48550/arXiv.2609.07954}
}

@article{flashsinkhorn2026,
  title={{FlashSinkhorn}: {IO}-Aware Entropic Optimal Transport on {GPU}},
  author={Ye, Felix X.-F. and Li, Xingjie and Yu, An and Chang, Ming-Ching and Chu, Linsong and Wertheimer, Davis},
  journal={arXiv preprint arXiv:2602.03067},
  year={2026},
  doi={10.48550/arXiv.2602.03067}
}

@inproceedings{vaswani2017attention,
  title={Attention Is All You Need},
  author={Vaswani, Ashish and Shazeer, Noam and Parmar, Niki and Uszkoreit, Jakob and Jones, Llion and Gomez, Aidan N. and Kaiser, {\L}ukasz and Polosukhin, Illia},
  booktitle={Advances in Neural Information Processing Systems},
  volume={30},
  year={2017}
}

@inproceedings{dao2022flashattention,
  title={{FlashAttention}: Fast and Memory-Efficient Exact Attention with {IO}-Awareness},
  author={Dao, Tri and Fu, Daniel Y. and Ermon, Stefano and Rudra, Atri and R{\'e}, Christopher},
  booktitle={Advances in Neural Information Processing Systems},
  volume={35},
  year={2022}
}

@inproceedings{dao2023flashattention2,
  title={{FlashAttention-2}: Faster Attention with Better Parallelism and Work Partitioning},
  author={Dao, Tri},
  booktitle={International Conference on Learning Representations},
  year={2024}
}

@inproceedings{shah2024flashattention3,
  title={{FlashAttention-3}: Fast and Accurate Attention with Asynchrony and Low-Precision},
  author={Shah, Jay and Bikshandi, Ganesh and Zhang, Ying and Thakkar, Vijay and Ramani, Pradeep and Dao, Tri},
  booktitle={Advances in Neural Information Processing Systems},
  volume={37},
  year={2024}
}

@article{beltagy2020longformer,
  title={Longformer: The Long-Document Transformer},
  author={Beltagy, Iz and Peters, Matthew E. and Cohan, Arman},
  journal={arXiv preprint arXiv:2004.05150},
  year={2020},
  doi={10.48550/arXiv.2004.05150}
}

@inproceedings{zaheer2020bigbird,
  title={Big Bird: Transformers for Longer Sequences},
  author={Zaheer, Manzil and Guruganesh, Guru and Dubey, Kumar Avinava and Ainslie, Joshua and Alberti, Chris and Onta{\~n}{\'o}n, Santiago and Pham, Philip and Ravula, Anirudh and Wang, Qifan and Yang, Li and Ahmed, Amr},
  booktitle={Advances in Neural Information Processing Systems},
  volume={33},
  pages={17283--17297},
  year={2020}
}

@inproceedings{jaegle2021perceiver,
  title={Perceiver: General Perception with Iterative Attention},
  author={Jaegle, Andrew and Gimeno, Felix and Brock, Andy and Vinyals, Oriol and Zisserman, Andrew and Carreira, Jo{\~a}o},
  booktitle={Proceedings of the 38th International Conference on Machine Learning},
  series={Proceedings of Machine Learning Research},
  volume={139},
  pages={4651--4664},
  year={2021}
}

@inproceedings{darcet2024vision,
  title={Vision Transformers Need Registers},
  author={Darcet, Timoth{\'e}e and Oquab, Maxime and Mairal, Julien and Bojanowski, Piotr},
  booktitle={International Conference on Learning Representations},
  year={2024}
}

@inproceedings{tay2020sparse,
  title={Sparse {Sinkhorn} Attention},
  author={Tay, Yi and Bahri, Dara and Yang, Liu and Metzler, Donald and Juan, Da-Cheng},
  booktitle={Proceedings of the 37th International Conference on Machine Learning},
  series={Proceedings of Machine Learning Research},
  volume={119},
  pages={9438--9447},
  year={2020}
}

@inproceedings{sander2022sinkformers,
  title={Sinkformers: Transformers with Doubly Stochastic Attention},
  author={Sander, Micha{\"e}l E. and Ablin, Pierre and Blondel, Mathieu and Peyr{\'e}, Gabriel},
  booktitle={Proceedings of the 25th International Conference on Artificial Intelligence and Statistics},
  series={Proceedings of Machine Learning Research},
  volume={151},
  pages={3515--3530},
  year={2022}
}

@article{shahbazi2025lotformer,
  title={{LOTFormer}: Doubly-Stochastic Linear Attention via Low-Rank Optimal Transport},
  author={Shahbazi, Ashkan and Thrash, Chayne and Bai, Yikun and Hamm, Keaton and NaderiAlizadeh, Navid and Kolouri, Soheil},
  journal={arXiv preprint arXiv:2509.23436},
  year={2025},
  doi={10.48550/arXiv.2509.23436}
}

@inproceedings{martins2016softmax,
  title={From Softmax to Sparsemax: A Sparse Model of Attention and Multi-Label Classification},
  author={Martins, Andr{\'e} F. T. and Astudillo, Ram{\'o}n Fernandez},
  booktitle={Proceedings of the 33rd International Conference on Machine Learning},
  series={Proceedings of Machine Learning Research},
  volume={48},
  pages={1614--1623},
  year={2016}
}

@inproceedings{correia2019adaptively,
  title={Adaptively Sparse Transformers},
  author={Correia, Gon{\c{c}}alo M. and Niculae, Vlad and Martins, Andr{\'e} F. T.},
  booktitle={Proceedings of the 2019 Conference on Empirical Methods in Natural Language Processing and the 9th International Joint Conference on Natural Language Processing (EMNLP-IJCNLP)},
  pages={2174--2184},
  year={2019}
}

@inproceedings{vasylenko2025longcontext,
  title={Long-Context Generalization with Sparse Attention},
  author={Vasylenko, Pavlo and Pitorro, Hugo and Martins, Andr{\'e} F. T. and Treviso, Marcos},
  booktitle={International Conference on Learning Representations},
  year={2026}
}

@inproceedings{velickovic2025softmax,
  title={Softmax is not Enough (for Sharp Size Generalisation)},
  author={Veli{\v{c}}kovi{\'c}, Petar and Perivolaropoulos, Christos and Barbero, Federico and Pascanu, Razvan},
  booktitle={Proceedings of the 42nd International Conference on Machine Learning},
  series={Proceedings of Machine Learning Research},
  volume={267},
  pages={61190--61211},
  year={2025}
}

@article{wortsman2023replacing,
  title={Replacing Softmax with {ReLU} in Vision Transformers},
  author={Wortsman, Mitchell and Lee, Jaehoon and Gilmer, Justin and Kornblith, Simon},
  journal={arXiv preprint arXiv:2309.08586},
  year={2023}
}

@inproceedings{cuturi2013sinkhorn,
  title={Sinkhorn Distances: Lightspeed Computation of Optimal Transport},
  author={Cuturi, Marco},
  booktitle={Advances in Neural Information Processing Systems},
  volume={26},
  year={2013}
}

@article{peyre2019computational,
  title={Computational Optimal Transport with Applications to Data Sciences},
  author={Peyr{\'e}, Gabriel and Cuturi, Marco},
  journal={Foundations and Trends in Machine Learning},
  volume={11},
  number={5--6},
  pages={355--607},
  year={2019},
  doi={10.1561/2200000073}
}

@inproceedings{blondel2018smooth,
  title={Smooth and Sparse Optimal Transport},
  author={Blondel, Mathieu and Seguy, Vivien and Rolet, Antoine},
  booktitle={Proceedings of the 21st International Conference on Artificial Intelligence and Statistics},
  series={Proceedings of Machine Learning Research},
  volume={84},
  pages={880--889},
  year={2018}
}

@inproceedings{sarlin2020superglue,
  title={{SuperGlue}: Learning Feature Matching with Graph Neural Networks},
  author={Sarlin, Paul-Edouard and DeTone, Daniel and Malisiewicz, Tomasz and Rabinovich, Andrew},
  booktitle={Proceedings of the IEEE/CVF Conference on Computer Vision and Pattern Recognition},
  year={2020}
}

@article{sinkhornknopp1967,
  title={Concerning Nonnegative Matrices and Doubly Stochastic Matrices},
  author={Sinkhorn, Richard and Knopp, Paul},
  journal={Pacific Journal of Mathematics},
  volume={21},
  number={2},
  pages={343--348},
  year={1967}
}

@article{knight2008sinkhorn,
  title={The {Sinkhorn--Knopp} Algorithm: Convergence and Applications},
  author={Knight, Philip A.},
  journal={SIAM Journal on Matrix Analysis and Applications},
  volume={30},
  number={1},
  pages={261--275},
  year={2008},
  doi={10.1137/060659624}
}

@article{chizat2018scaling,
  title={Scaling Algorithms for Unbalanced Optimal Transport Problems},
  author={Chizat, L{\'e}na{\"\i}c and Peyr{\'e}, Gabriel and Schmitzer, Bernhard and Vialard, Fran{\c{c}}ois-Xavier},
  journal={Mathematics of Computation},
  volume={87},
  number={314},
  pages={2563--2609},
  year={2018}
}

@inproceedings{luise2018differential,
  title={Differential Properties of {Sinkhorn} Approximation for Learning with {Wasserstein} Distance},
  author={Luise, Giulia and Rudi, Alessandro and Pontil, Massimiliano and Ciliberto, Carlo},
  booktitle={Advances in Neural Information Processing Systems},
  volume={31},
  pages={5864--5874},
  year={2018}
}

@inproceedings{eisenberger2022unified,
  title={A Unified Framework for Implicit {Sinkhorn} Differentiation},
  author={Eisenberger, Marvin and Toker, Aysim and Leal-Taix{\'e}, Laura and Bernard, Florian and Cremers, Daniel},
  booktitle={Proceedings of the IEEE/CVF Conference on Computer Vision and Pattern Recognition},
  pages={509--518},
  year={2022}
}

@inproceedings{thornton2023initialization,
  title={Rethinking Initialization of the {Sinkhorn} Algorithm},
  author={Thornton, James and Cuturi, Marco},
  booktitle={Proceedings of the 26th International Conference on Artificial Intelligence and Statistics},
  series={Proceedings of Machine Learning Research},
  volume={206},
  pages={8682--8698},
  year={2023}
}

@inproceedings{merity2017pointer,
  title={Pointer Sentinel Mixture Models},
  author={Merity, Stephen and Xiong, Caiming and Bradbury, James and Socher, Richard},
  booktitle={International Conference on Learning Representations},
  year={2017}
}

@article{su2024roformer,
  title={{RoFormer}: Enhanced Transformer with Rotary Position Embedding},
  author={Su, Jianlin and Ahmed, Murtadha and Lu, Yu and Pan, Shengfeng and Bo, Wen and Liu, Yunfeng},
  journal={Neurocomputing},
  volume={568},
  pages={127063},
  year={2024}
}

@inproceedings{press2022train,
  title={Train Short, Test Long: Attention with Linear Biases Enables Input Length Extrapolation},
  author={Press, Ofir and Smith, Noah A. and Lewis, Mike},
  booktitle={International Conference on Learning Representations},
  year={2022}
}

@inproceedings{peng2024yarn,
  title={{YaRN}: Efficient Context Window Extension of Large Language Models},
  author={Peng, Bowen and Quesnelle, Jeffrey and Fan, Honglu and Shippole, Enrico},
  booktitle={International Conference on Learning Representations},
  year={2024}
}

@inproceedings{xiao2024efficient,
  title={Efficient Streaming Language Models with Attention Sinks},
  author={Xiao, Guangxuan and Tian, Yuandong and Chen, Beidi and Han, Song and Lewis, Mike},
  booktitle={International Conference on Learning Representations},
  year={2024}
}

@article{liu2024lost,
  title={Lost in the Middle: How Language Models Use Long Contexts},
  author={Liu, Nelson F. and Lin, Kevin and Hewitt, John and Paranjape, Ashwin and Bevilacqua, Michele and Petroni, Fabio and Liang, Percy},
  journal={Transactions of the Association for Computational Linguistics},
  volume={12},
  pages={157--173},
  year={2024}
}

@inproceedings{hsieh2024ruler,
  title={{RULER}: What's the Real Context Size of Your Long-Context Language Models?},
  author={Hsieh, Cheng-Ping and Sun, Simeng and Kriman, Samuel and Acharya, Shantanu and Rekesh, Dima and Jia, Fei and Zhang, Yang and Ginsburg, Boris},
  booktitle={Conference on Language Modeling},
  year={2024}
}

@inproceedings{arora2024zoology,
  title={Zoology: Measuring and Improving Recall in Efficient Language Models},
  author={Arora, Simran and Eyuboglu, Sabri and Timalsina, Aman and Johnson, Isys and Poli, Michael and Zou, James and Rudra, Atri and R{\'e}, Christopher},
  booktitle={International Conference on Learning Representations},
  year={2024}
}

@inproceedings{katharopoulos2020transformers,
  title={Transformers are {RNN}s: Fast Autoregressive Transformers with Linear Attention},
  author={Katharopoulos, Angelos and Vyas, Apoorv and Pappas, Nikolaos and Fleuret, Fran{\c{c}}ois},
  booktitle={Proceedings of the 37th International Conference on Machine Learning},
  series={Proceedings of Machine Learning Research},
  volume={119},
  pages={5156--5165},
  year={2020}
}

@inproceedings{gu2024mamba,
  title={Mamba: Linear-Time Sequence Modeling with Selective State Spaces},
  author={Gu, Albert and Dao, Tri},
  booktitle={Conference on Language Modeling},
  year={2024}
}

@inproceedings{yang2024gla,
  title={Gated Linear Attention Transformers with Hardware-Efficient Training},
  author={Yang, Songlin and Wang, Bailin and Shen, Yikang and Panda, Rameswar and Kim, Yoon},
  booktitle={Proceedings of the 41st International Conference on Machine Learning},
  series={Proceedings of Machine Learning Research},
  volume={235},
  pages={56501--56523},
  year={2024}
}

@inproceedings{zhang2024gsa,
  title={Gated Slot Attention for Efficient Linear-Time Sequence Modeling},
  author={Zhang, Yu and Yang, Songlin and Zhu, Ruijie and Zhang, Yue and Cui, Leyang and Wang, Yiqiao and Wang, Bolun and Shi, Freda and Wang, Bailin and Bi, Wei and Zhou, Peng and Fu, Guohong},
  booktitle={Advances in Neural Information Processing Systems},
  volume={37},
  pages={116870--116898},
  year={2024}
}

@inproceedings{triton2019tillet,
  title={Triton: An Intermediate Language and Compiler for Tiled Neural Network Computations},
  author={Tillet, Philippe and Kung, H. T. and Cox, David},
  booktitle={Proceedings of the 3rd ACM SIGPLAN International Workshop on Machine Learning and Programming Languages},
  pages={10--19},
  year={2019}
}
}

\end{document}